\documentclass[10pt,twocolumn,letterpaper]{article}

\usepackage[pagenumbers]{iccv} 
\usepackage{lineno}
\usepackage{booktabs}
\usepackage{arydshln}
\usepackage{bbm}
\usepackage{threeparttable}
\usepackage{color,graphicx}
\usepackage{amsmath}
\usepackage{mathrsfs}
\usepackage{amsthm}
\usepackage[ruled]{algorithm2e}
\usepackage{bm}
\usepackage{multirow}
\usepackage{amssymb}
\usepackage{bm}
\usepackage{multirow}
\usepackage{arydshln}
\usepackage{amssymb}
\usepackage{nicematrix,booktabs,caption}
\usepackage[ruled]{algorithm2e}

\usepackage[dvipsnames]{xcolor}
\usepackage{pifont}

\definecolor{iccvblue}{rgb}{0.21,0.49,0.74}
\usepackage[pagebackref,breaklinks,colorlinks,allcolors=iccvblue]{hyperref}

\def\paperID{8985} 
\def\confName{ICCV}
\def\confYear{2025}

\definecolor{babyblue}{rgb}{0.63, 0.79, 0.95}
\definecolor{babypink}{rgb}{0.96, 0.76, 0.76}
\title{ProCrop: Learning Aesthetic Image Cropping from Professional Compositions}

\author{
Ke Zhang$^{1}$ \quad
Tianyu Ding$^{3 \dagger}$ \quad
Jiachen Jiang$^{2}$ \quad
Tianyi Chen$^{3}$ \quad\\
Ilya Zharkov$^{3}$ \quad
Vishal M. Patel$^{1}$ \quad
Luming Liang$^{3\dagger}$ \\
\\
$^1$Johns Hopkins University \quad
$^2$Ohio State University \quad
$^3$Microsoft \\
\thanks{$\dagger$ Corresponding author.}
\url{https://bwgzk-keke.github.io/ProCrop/}
}

\begin{document}
\maketitle

\begin{abstract}
{Image cropping is crucial for enhancing the visual appeal and narrative impact of photographs, yet existing rule-based and data-driven approaches often lack diversity or require annotated training data. We introduce ProCrop, a retrieval-based method that leverages professional photography to guide cropping decisions. By fusing features from professional photographs with those of the query image, ProCrop learns from professional compositions, significantly boosting performance. Additionally, we present a large-scale dataset of 242K weakly-annotated images, generated by out-painting professional images and iteratively refining diverse crop proposals. This composition-aware dataset generation offers diverse high-quality crop proposals guided by aesthetic principles and becomes the largest publicly available dataset for image cropping. Extensive experiments show that ProCrop significantly outperforms existing methods in both supervised and weakly-supervised settings. Notably, when trained on the new dataset, our ProCrop surpasses previous weakly-supervised methods and even matches fully supervised approaches. Both the code and dataset will be made publicly available to advance research in image aesthetics and composition analysis.}


\end{abstract}    
\section{Introduction}
\label{sec:intro}

{In visual arts, a well-composed photograph can captivate viewers and convey profound messages. Image cropping, the art of selectively removing peripheral areas from a photograph, is crucial for enhancing visual appeal and narrative potency. However, achieving aesthetically pleasing compositions through cropping is challenging due to the intricate interplay of various compositional elements~\cite{liu2010optimizing,obrador2010role}, especially for non-professionals and automated systems. 
}

{Existing automatic image cropping methods typically fall into two categories: those guided by composition rules in photography~\cite{fang2014automatic,ni2013learning,zhang2013weakly} and data-driven approaches such as anchor-based~\cite{li2020composing,lian2021context,zeng2019reliable,zeng2020grid,wang2023image} and coordinate regression-based~\cite{guo2018automatic,hong2021composing,li2020learning,liu2023beyond} methods. Rule-based approaches often struggle to fully capture sophisticated features and complex compositions, being constrained by the very principles they're founded upon. Data-driven methods, while promising, face challenges due to their reliance on annotated datasets for training. Creating large-scale, diverse datasets of aesthetically pleasing compositions is labor-intensive and time-consuming.  Currently, the largest available dataset for this task contains only about 10K images (see \cref{tab:data}), which is insufficient to capture the vast diversity of compositions and styles found in professional photography.}

\begin{figure}[!t]
\centering
         \includegraphics[width=\linewidth]{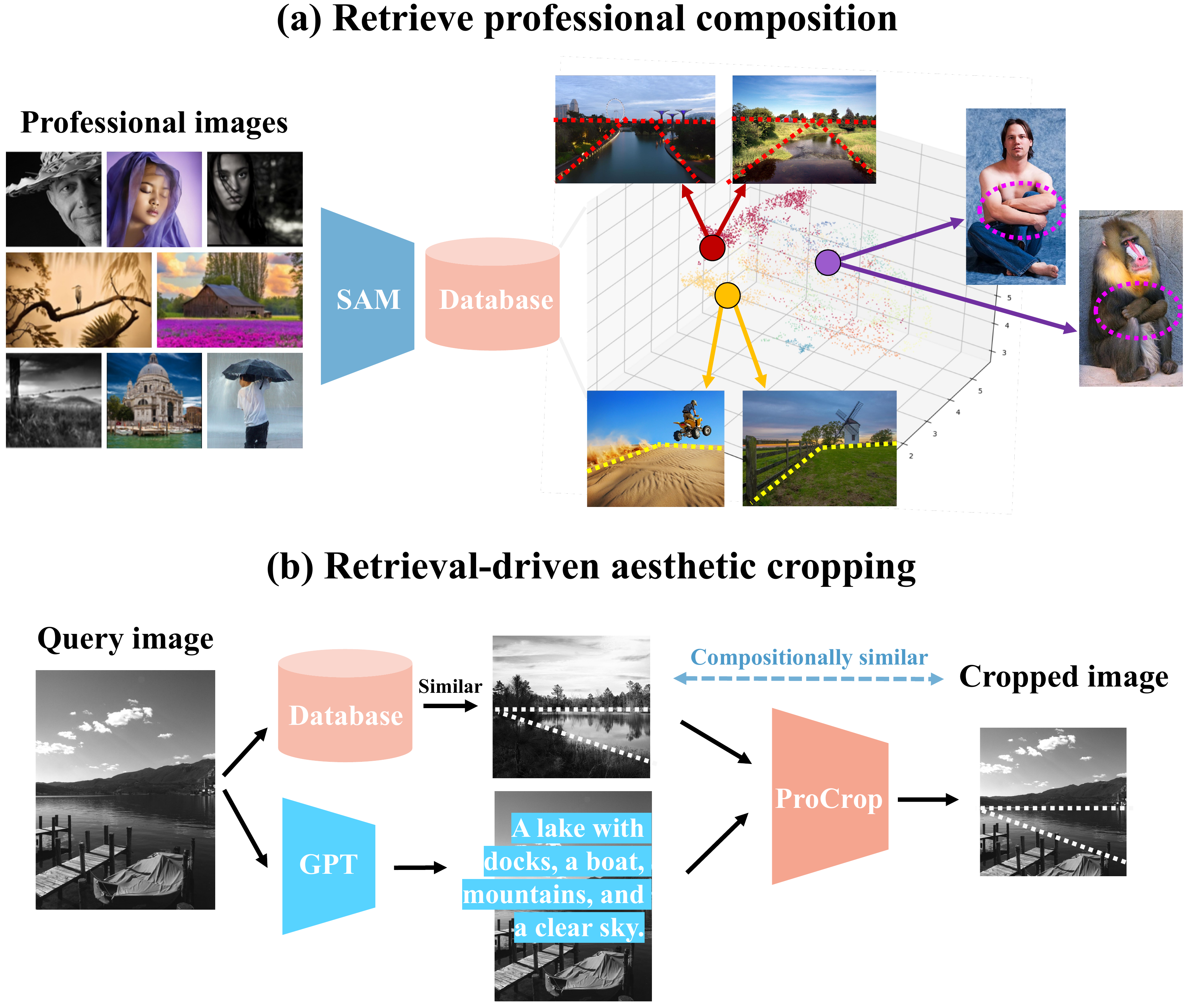}
\caption{
{Overview of ProCrop's retrieval-based aesthetic cropping approach. (a) Construction of a professional image database and retrieval of images with similar compositional layouts. (b) Demonstration of ProCrop's cropping process, where a compositionally similar reference image guides the generation of aesthetically pleasing crop results.}
}
\label{fig:teaser}
\vspace{-.1in}
\end{figure}

{In this paper, we introduce a novel retrieval-based image cropping approach that harnesses the wealth of existing professional photography. Inspired by retrieval augmentation in language models~\cite{borgeaud2022improving,guu2020retrieval} and the abundance of professional photography datasets, we learn from professional images with similar aesthetic compositions~(see~\cref{fig:teaser}). Our key insight is that professional photographers have already solved numerous compositional challenges through their experience and artistic vision. By tapping into this knowledge base, we guide our model to align with professional standards. This approach addresses diversity limitations in rule-based methods while enhancing data-driven methods with external knowledge. Importantly, this requires no annotations for the reference database, ensuring its practicality. We demonstrate that integrating this retrieval-augmented concept into image cropping yields state-of-the-art (SOTA) performance, underscoring its effectiveness.
}

{Furthermore, we address the scarcity of high-quality aesthetic training data by developing a large-scale dataset through a weakly-supervised approach. Specifically, we leverage ControlNet~\cite{zhang2023adding}, a text-to-image diffusion model, to outpaint professional images, simulating cropped and uncropped pairs. Starting with AVA~\cite{6247954} and unsplash-lite~\cite{unsplash2023}, the large collection of professional images serving as expert labels (\ie, good crops), we employ GPT-4~\cite{achiam2023gpt} to infer textual layouts beyond original image boundaries and use SAM~\cite{kirillov2023segment} to extract multi-scale compositional masks. These are then fed into ControlNet for image outpainting. Through an iterative refinement process, we generate diverse crop proposals, substantially expanding the available data. The resulting dataset comprises 242K annotated aesthetic images, significantly surpassing existing resources in scale and diversity~(see~\cref{tab:data}). Our weakly supervised training on this dataset, combined with image retrieval, not only outperforms previous weakly supervised methods but also achieves results comparable to fully supervised ones. 
}


{
Our contributions are summarized as follows:
\begin{itemize}
\item We propose ProCrop, a retrieval-based image cropping method that leverages professional photography knowledge to achieve aesthetically pleasing compositions.
\item We introduce a new dataset through a weakly-supervised, controlled approach. To the best of our knowledge, this is the largest dataset for aesthetic image cropping.
\item Experiments show that our retrieval-based method significantly outperforms existing works. Notably, trained on our new dataset, it surpasses prior weakly-supervised methods and even matches fully supervised approaches.
\end{itemize}
We will make both the code and dataset publicly available. This large-scale dataset is expected to enhance image cropping techniques and serve as a valuable resource for the broader computer vision community, advancing research in image aesthetics and composition analysis.
}

\section{Related work}

\subsection{Aesthetic image cropping}

{Aesthetics image cropping aims to enhance the visual appeal of images by learning aesthetic composition via comparative views. Unlike related tasks such as image retargeting~\cite{rubinstein2010comparative, setlur2005automatic} that primarily focus on content preservation, aesthetic cropping typically generates candidate crops via scaling and shifting, and scoring them based on aesthetics.}

{Image cropping methods can be broadly categorized into rule-based and data-driven approaches. Rule-based methods~\cite{hong2021composing,fang2014automatic,ni2013learning,zhang2013weakly} rely on hand-crafted features and techniques like saliency detection~\cite{vig2014large} or specific aesthetic rules~\cite{liu2010optimizing,nishiyama2009sensation,zhang2005auto}. While effective at content preservation, they often struggle with nuanced compositions. Data-driven approaches, which now dominate the field, include anchor-based methods~\cite{li2020composing,chen2017learning,tu2020image,zeng2019reliable,wei2018good,wang2017deep,lian2021context,zeng2020grid} that evaluate candidate regions, and coordinate regression methods~\cite{chen2017learning,li2018a2,guo2018automatic,hong2021composing,li2020learning,liu2023beyond} that directly predict crop boundaries. In contrast to these existing approaches, our proposed retrieval-based method offers a novel perspective. By leveraging a large corpus of professional images, our approach overcomes the limitations of hand-crafted rules and the need for extensive labeled datasets, enabling more flexible and context-aware cropping decisions.}

{A crucial aspect of data-driven methods is their dependence on large-scale supervised training. Widely used datasets such as GAICv1~\cite{zeng2019reliable}, GAICv2~\cite{zeng2020grid}, CPC~\cite{wei2018good}, FCDB~\cite{chen2017quantitative}, and SACD~\cite{yang2023focusing} are labor-intensive and expensive to create. Recently, \cite{hong2024learning} attempted to address these issues by outpainting professional images. However, their approach is constrained to single crop suggestions, faces reliability issues with out-painted content, and is not publicly accessible. In this work, we present a large-scale dataset of weakly-annotated images, generated by out-painting professional images and iteratively refining diverse crop proposals. Our composition-aware approach yields high-quality and diverse crop proposals. By making this dataset publicly available, we aim to advance research in the field of image cropping and composition.}

\begin{table}[t]
\centering
\caption{Summary of datasets for image cropping.}\label{tab:data}
\vspace{-.1in}
\resizebox{\linewidth}{!}{
\begin{NiceTabular}{l|c c| c c c}
\CodeBefore
\rowcolor{babyblue!30}{11}
\Body
\toprule
\multirow{2}{*}{Datasets}  & \multirow{2}{*}{Year} & \multirow{2}{*}{Venue} & \multirow{2}{*}{\# of Images} & \multicolumn{2}{c}{\# of Annotations}\\
\cmidrule{5-6}
&&&&Avg&Total\\
\midrule
ICDB~\cite{yan2013learning}  & 2013 & CVPR                  & 1,000  & 1 & 1000 \\
FLMS~\cite{fang2014automatic} & 2014 & ACM                   & 500   & 10 & 5000\\
FCDB~\cite{chen-wacv2017} & 2017 & WACV                  & 1,743  & 1 &  1743\\
CPC~\cite{wei2018good} & 2018 & CVPR                  & 10,797 & 24 & 259,128\\
GAICv1~\cite{zeng2019reliable} & 2019 & CVPR                  & 1,036  & 90& 93,240\\
GAICv2~\cite{zeng2020grid} & 2020 & TPAMI & 2,626  & 90 & 236,340\\
SACD~\cite{yang2023focusing} & 2023 & CVM & 2,777 & 8  &22,216\\
UGCrop5K~\cite{ko2024semantic} & 2024 & AAAI & 5,000  & 90 & 450,000\\
\textbf{Ours} & 2025 & Under review &242,000& 8&1,936,000\\
\bottomrule
\end{NiceTabular}}
\vspace{-.1in}
\end{table}

\begin{figure*}[!t]
\centering
         \includegraphics[width=\textwidth]{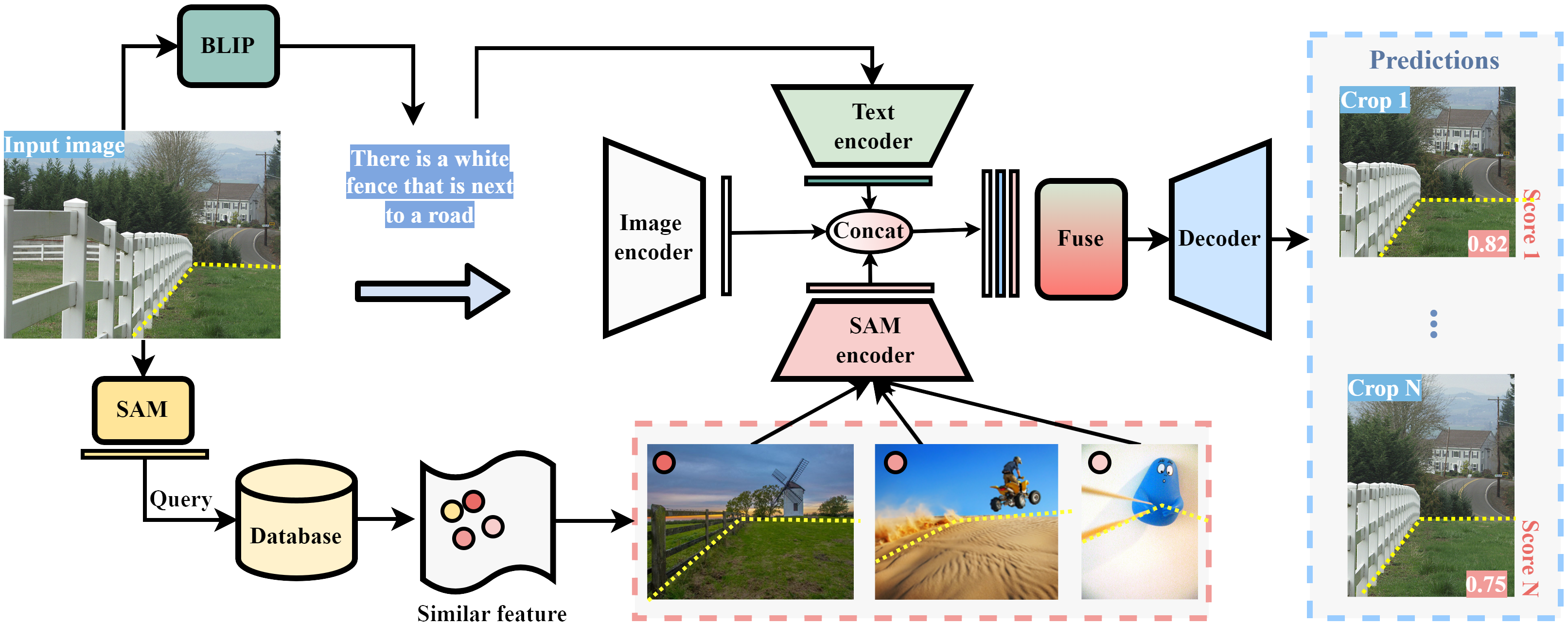}
\caption{
{The pipeline of ProCrop. Given an input image, ProCrop retrieves compositionally similar professional images and generates a textual description, which guide the model to produce aesthetically enhanced crops along with corresponding aesthetic scores.}
}
\label{fig:figure2}
\end{figure*}

\subsection{Retrieval augmentation}

{Retrieval augmentation~\cite{asai2023retrieval,blattmann2022retrieval,borgeaud2022improving,guu2020retrieval,qin2022highly} provides an effective approach to improve model performance without expanding model parameters or requiring additional training data. Instead of storing all knowledge within model parameters, these techniques utilize external database to fetch relevant information on demand.
A typical method is to fetch $k$-nearest neighbors from a pre-computed embedding space to provide supplementary input.
This strategy has demonstrated success across various domains, including language models~\cite{guu2020retrieval}, diffusion models~\cite{qin2022highly}, and layout generation~\cite{horita2024retrieval}. In composition-aware image cropping, a fundamental challenge is to effectively encode both visual content and aesthetic rules. While previous works often struggle with data scarcity~\cite{qian2020retrieve}, our retrieval-based framework leverages existing professional images, enabling the model to learn and apply sophisticated composition rules while maintaining computational efficiency.}

\section{Method}

\subsection{Overview}

{Image cropping aims to enhance the composition of photographs that may not have been captured professionally. Given an input image $I\in\mathbb{R}^{H\times W\times 3}$, our goal is to predict a series of crop rectangles with high aesthetic scores, denoted as $\{(\bm b_n, s_n)\}_{n=1}^N$, where $\bm b_n\in[0,1]^4$ is the bounding box in normalized coordinates, $s_n$ is the  aesthetic quality, and $N$ is the number of predicted crops. We use the image as input and its ground truth crops or pseudo labels for supervision. This task presents significant challenges due to the intricate interplay of various compositional elements, such as subject positioning, adherence to the rule of thirds, and the use of leading lines.}

{Inspired by how retrieval augmentation has improved the quality of language models and image synthesis, we propose a novel module for retrieval-based aesthetic image cropping (\cref{sec:propcrop}). Our approach learns from professional compositions without requiring additional annotations on the retrieval database, significantly improving the quality of generated compositions. Furthermore, we introduce a composition-aware approach to generate a large dataset (\cref{sec:data}). This method offers multiple high-quality crop proposals guided by aesthetic principles, enhancing the learning process and ultimate performance of our model.}


\subsection{ProCrop: Retrieval-driven aesthetic cropping}\label{sec:propcrop}

{To effectively leverage professional images, we introduce a retrieval module that addresses two key challenges: (1) retrieving reference images from a database based on their compositional features, and (2) fusing the retrieved features into a final augmented representation. 
Our approach is inspired by the assumption that compositional features can be characterized by line combinations in images~\cite{lee2018photographic,ko2024semantic}. To capture these line compositions in retrieved images, we employ SAM~\cite{kirillov2023segment}, which offers richer, more precise boundaries without relying on direct semantic mask extraction, compared to CLIP~\cite{radford2021learning} or saliency map~\cite{hoh2023salient,zhang2018detecting,horita2024retrieval}.  Appendix D  presents more details.}

{\textbf{Feature retrieval.} Let $\mathcal{V}$ represent the database of professional images. For an input image $I$, we aim to identify professional images with the most similar compositional characteristics. We use the SAM encoder to extract features from both the query image $I$ and each image in the professional database $\mathcal{V}$, yielding $f_I$ and $\bm F=\{f_{\widetilde{I}} \ | \ \widetilde{I}\in\mathcal{V}\}$, respectively. Based on feature similarity, we retrieve the top-$K$ most similar compositional features in $\bm F$, represented by $\bm R\in\mathbb{R}^{K\times m\times d}$, where $m$ is the flattened spatial dimension and $d$ is the feature dimension. To streamline the training process, we precompute and cache the SAM feature embeddings for each training image along with their $K$ most similar counterparts from $\mathcal{V}$. These image-embedding pairs are indexed in a database, with ElasticSearch~\cite{ElasticSearch_Huggingface} serving as the retrieval engine for efficient similarity-based matching.}

\begin{figure*}[!t]
\centering
        \includegraphics[width=0.9\textwidth]{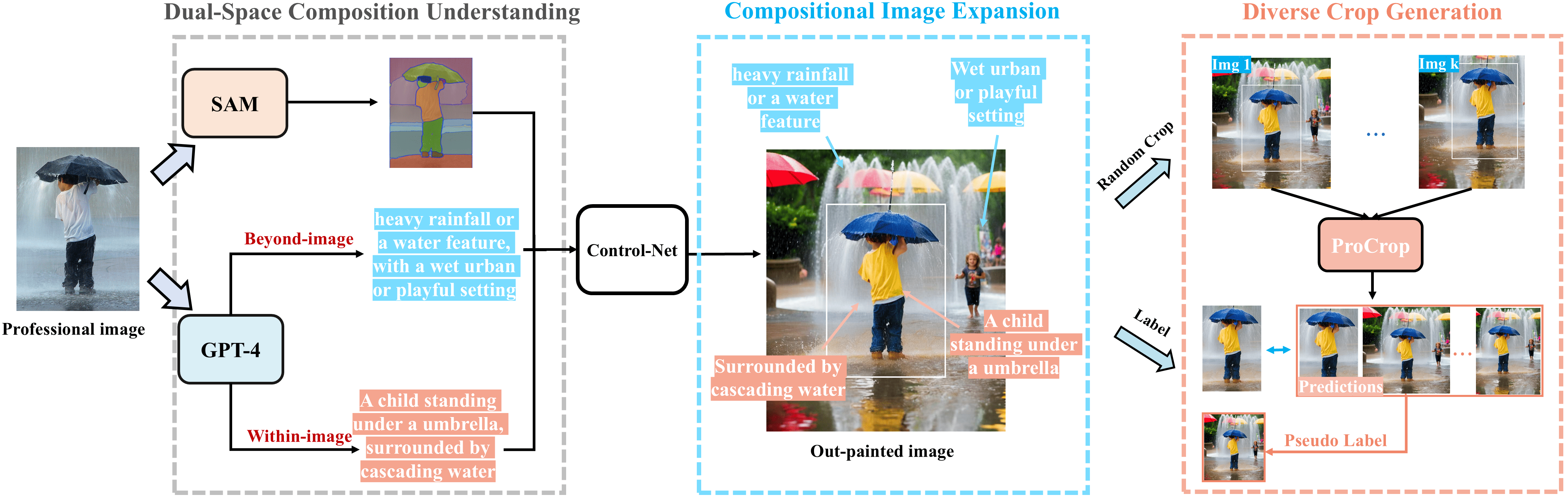}
\caption{
{Composition-aware dataset generation. Professional images undergo three stages to create diverse
image-crop pairs.}
}
\label{fig:data-gen}
\end{figure*}

{In contrast to existing retrieval methods~\cite{fu2024dreamsim,horita2024retrieval}  that  primarily emphasize category similarity, our method is specifically designed to capture compositional characteristics. Through the utilization of SAM embeddings and our streamlined retrieval pipeline, we effectively learn compositional knowledge from professional photographs.}

{\textbf{Feature fusion.} Given the retrieved top-$K$ image features $\bm R$ from $\mathcal{V}$, we fuse them with the query image's embedded features to guide the cropping process. While directly utilizing the SAM embedding $f_I$ is feasible, SAM's computational overhead leads to slow training and inference. Instead, we adopt an encoder architecture similar to Conditional DETR (cDETR)~\cite{meng2021conditional}, which offers superior training convergence and inference efficiency while maintaining comparable performance. We denote this query image feature as $\bar{f}_I\in\mathbb{R}^{p\times d}$, where $p$ represents the flattened spatial dimension specific to this encoder. To effectively fuse $\bar f_I$ with $\bm R$, we employ a learnable projection head $\Pi(\cdot)$ that transforms $\bm R$ to match the spatial-channel dimensions of $\bar f_I$. The final feature fusion is achieved through:
\begin{equation}
f_R = \text{Concat}(\bar f_I, \Pi(\bm R), f_c),
\end{equation}
where $f_c$ denotes the cross-attended feature obtained by using $\bar f_I$ as the query and $\bm R$ as both key and value. The resulting fused feature $f_R$ is subsequently fed into the rest of the pipeline, incorporating compositional knowledge retrieved from professional photography.}

{Motivated by the natural ability of language to highlight salient image regions, we enhance the model by integrating multi-modal features with the fused image features. For an input image $I$, we first employ BLIP~\cite{li2022blip} to generate compositional text descriptions that explicitly capture the desired objects and their spatial arrangements. We then leverage BLIP to extract multi-modal embeddings $\bm M \in\mathbb{R}^{m'\times d}$ from these image-text pairs, where $m'$ is the flattened spatial dimension specific to the BLIP encoder. We precompute this process for all training images. The multi-modal feature fusion is then computed as:
\begin{equation}
f_M = \text{Concat}(\bar f_{I}, \Pi'(\bm M), f'_c),
\end{equation}
where $\Pi'(\cdot)$ denotes a learnable projection head for harmonizing the feature dimensions, and $f'_c$ represents the cross-attended feature derived by utilizing $\bar f_I$ as the query and $\bm M$ as both key and value. Details on text embeddings are provided in Appendix C of the Supplementary Material.}

{We concatenate the multi-modal feature $f_M$ with the retrieved feature $f_R$ and feed this combined representation into a transformer decoder. Following~\cite{jia2022rethinking}, our decoder processes both the input features and learnable anchors through parallel regression and classification heads. This architecture generates $N$ crop proposals, each accompanied by its aesthetic score, which can be expressed as:
\begin{equation}
\text{Decoder}(f_R, f_M) \mapsto \{(\bm b_n, s_n)\}_{n=1}^N.
\end{equation}
\subsection{Composition-aware dataset generation}\label{sec:data}
Data-driven image cropping rely on annotated datasets for training. However, high-quality datasets containing images and their aesthetic crops are scarce due to the labor-intensive nature. To address this, we develop an automated pipeline for generating large-scale cropping datasets in a weakly-supervised manner, as shown in~\cref{fig:data-gen}. Our dataset encompasses diverse image categories, professional crop proposals, and compositional descriptions. The pipeline leverages quality-validated professional photographs from public sources. We employ language and segmentation foundation models to encode compositions both within and beyond image boundaries. These are then fed into a text-to-image diffusion model to generate outpainted images, simulating uncropped and cropped image pairs. More illustrations are provided in Appendix B.1 .}

\begin{figure}[!t]
\centering
         \includegraphics[width=0.9\linewidth]{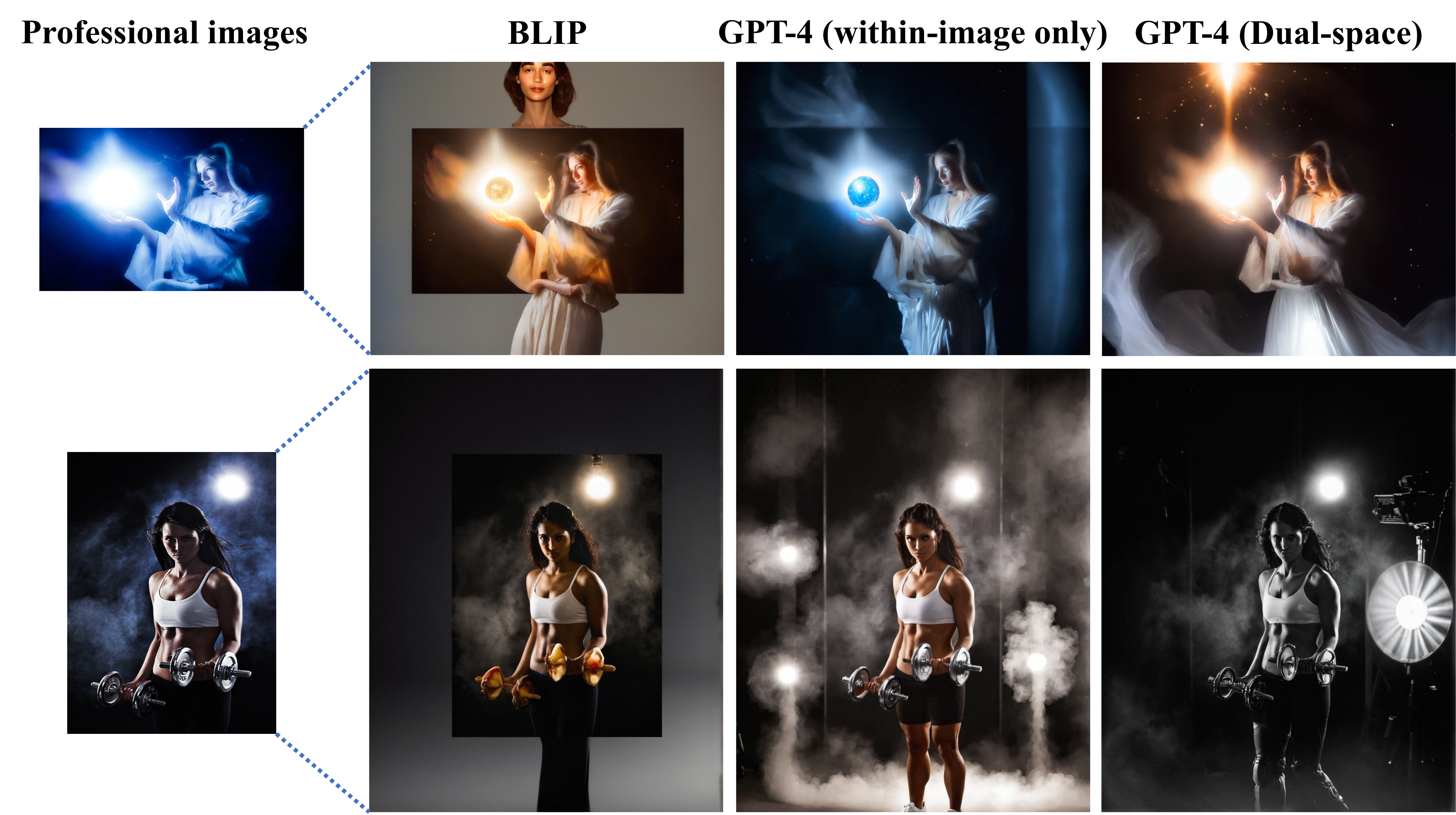}
\caption{
{Outpainting results with three variations: (1) BLIP-based composition understanding, (2) GPT-4 with solely within-image compositional descriptions, and (3) GPT-4 with the proposed dual-space composition understanding. The results show that our dual-space approach, through GPT-4, yields significantly more coherent and visually realistic outpainting outcomes.}
}
\label{fig:dual}
\end{figure}

{\textbf{Dual-space composition understanding.}
For \emph{within-image} compositions, we prompt GPT-4 to analyze compositional elements and identify salient subjects that attract human attention. We incorporate SAM-generated segmentation masks to ensure semantic consistency between input and generated content. For \emph{beyond-image} compositions, GPT-4 predicts potential content outside image boundaries and describes the broader context. Our experiments show that these beyond-image compositional descriptions are essential for effective outpainting, as shown in~\cref{fig:dual}. While~\cite{hong2024learning} proposes an outpainting approach for weakly annotated data, their reliance on image captions leads to artifacts like extraneous objects or unnatural grid patterns. In contrast, our composition-aware prompting strategy generates more coherent and visually plausible results.}


\begin{figure}[!t]
\centering
    \includegraphics[width=0.9\linewidth]{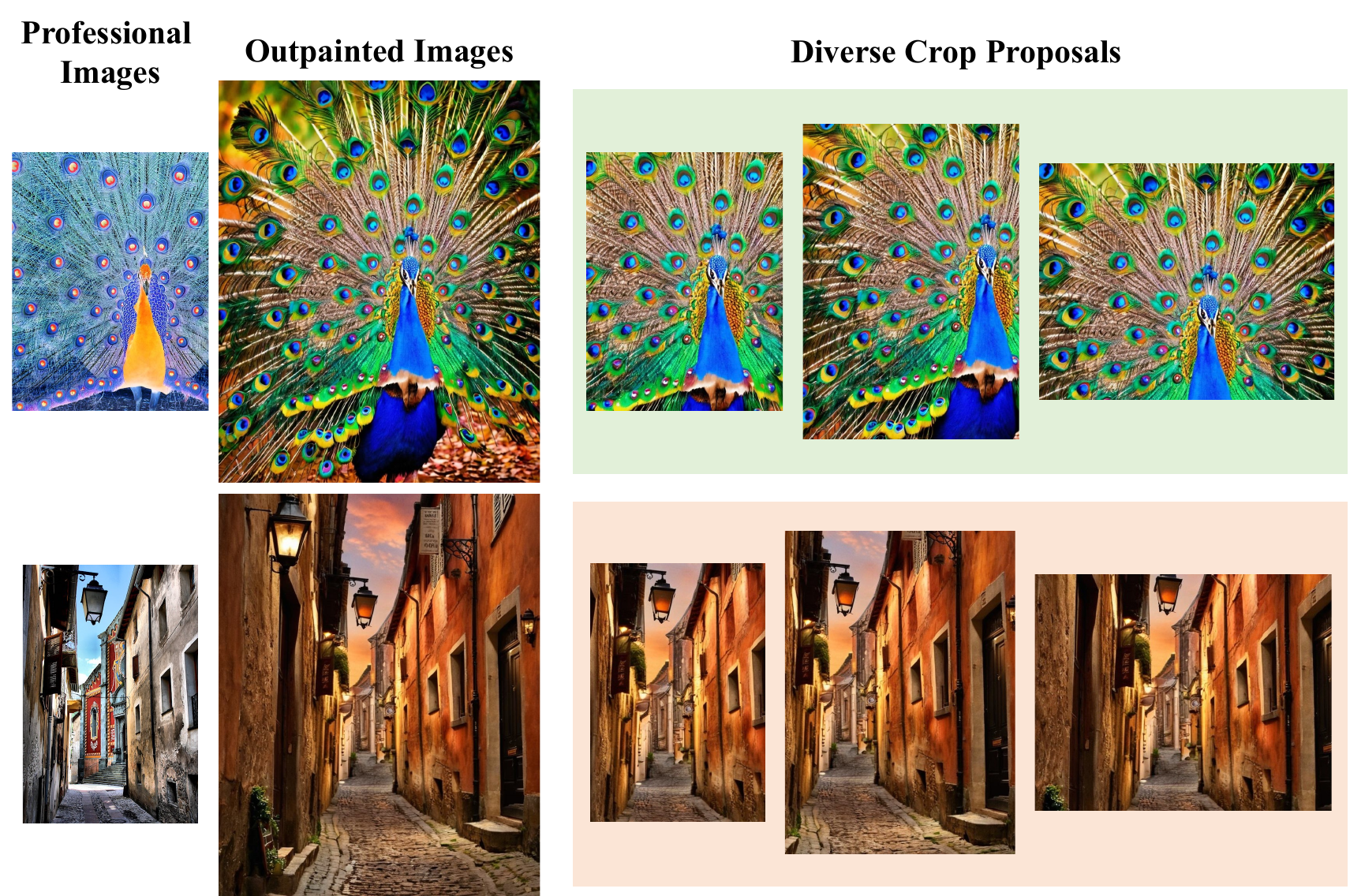}
\caption{
{Examples of outpainting results and crop proposals. Multiple crop proposals serve as high-quality pseudo-labels generated through the model-in-the-loop process.}
}
\label{fig:crop}
\end{figure}

{\textbf{Compositional image expansion.}
We randomly downscale the professional image and enlarge it to create a canvas with dimensions between 700 and 1024 pixels. The outpainting process feeds the canvas $I_c$, GPT-4 generated text descriptions $T$, and multi-scale SAM masks $S$ into a pretrained ControlNet~\cite{zhang2023adding} to produce the output $I'$:
\begin{equation}
I' = \text{ControlNet}(I_c, S, T).
\end{equation}
\indent\textbf{Diverse crop generation.}
Instead of the single crop proposal naturally arising from the original and outpainted images, we develop an iterative refinement process that creates high-quality, varied crop proposals (see~\cref{fig:crop}) through a model-in-the-loop approach. We generate random crops from expanded images, ensuring the preservation of original content  while varying in size and aspect ratio. These random crops serve as initial training inputs, with their corresponding original image regions acting as labels. We train a ProCrop model using these image-crop pairs. The model then enters an iterative cycle where it automatically generates crop proposals for each query image. These proposals undergo a curation process that selects a diverse set adhering to established aesthetic principles. During this iterative refinement process, we dynamically rank the aesthetic scores of the crop set. The top-$k$ crops are then utilized as pseudo labels, significantly enhancing the diversity of our crop annotations and ultimately improving the model's ability to generalize across various cropping scenarios.}

\section{Experiments}

\subsection{Datasets}

{We detail the datasets for image retrieval and cropping. More details are provided in Appendix B.}

{\textbf{Retrieval datasets.} We employ two datasets for image retrieval: CGL \cite{zhou2022composition} and AVA \cite{6247954}. 
CGL consists of 60,548 e-commerce posters, primarily featuring cosmetics and clothing advertisements with relatively simple compositional layouts. On the other hand, AVA is a significantly larger dataset containing 255,000 images with more complex scenarios and diverse compositional arrangements. From AVA, we select the top 55,000 images based on aesthetic scores to form the professional retrieval set.}

{\textbf{Cropping datasets.} We utilize five datasets for image cropping: GAICv1~\cite{zeng2019reliable}, GAICv2~\cite{zeng2020grid}, CPC~\cite{wei2018good}, FLMS~\cite{fang2014automatic}, and SACD~\cite{yang2023focusing}. GAIC and CPC serve as small and mid-sized training datasets, respectively, while SACD and FLMS are used for evaluation in zero-shot transfer experiments. The GAICv1 dataset contains 1,036 training and 200 testing images, with each image offering up to 90 crop proposals generated using a predefined grid-anchor system. GAICv2 is an extended version, consisting of 2,636 training images, 200 validation images, and 500 testing images. These proposals are rated on a 1-5 scale by six annotators through a two-stage process and organized into four aspect ratio groups, each containing six crops. The CPC dataset is a larger collection of 10,797 images, serving as a mid-sized benchmark for training supervised image cropping models. The FLMS dataset consists of 500 images, each accompanied by up to 10 high-quality crop proposals, and is exclusively used for testing purposes. Following~\cite{hong2024learning}, we utilize the test set of SACD for evaluation, which provides six to eight annotated cropping windows per image, focusing on aesthetic quality to ensure well-composed subjects.}


\begin{figure}[!t]
\centering         \includegraphics[width=0.9\linewidth]{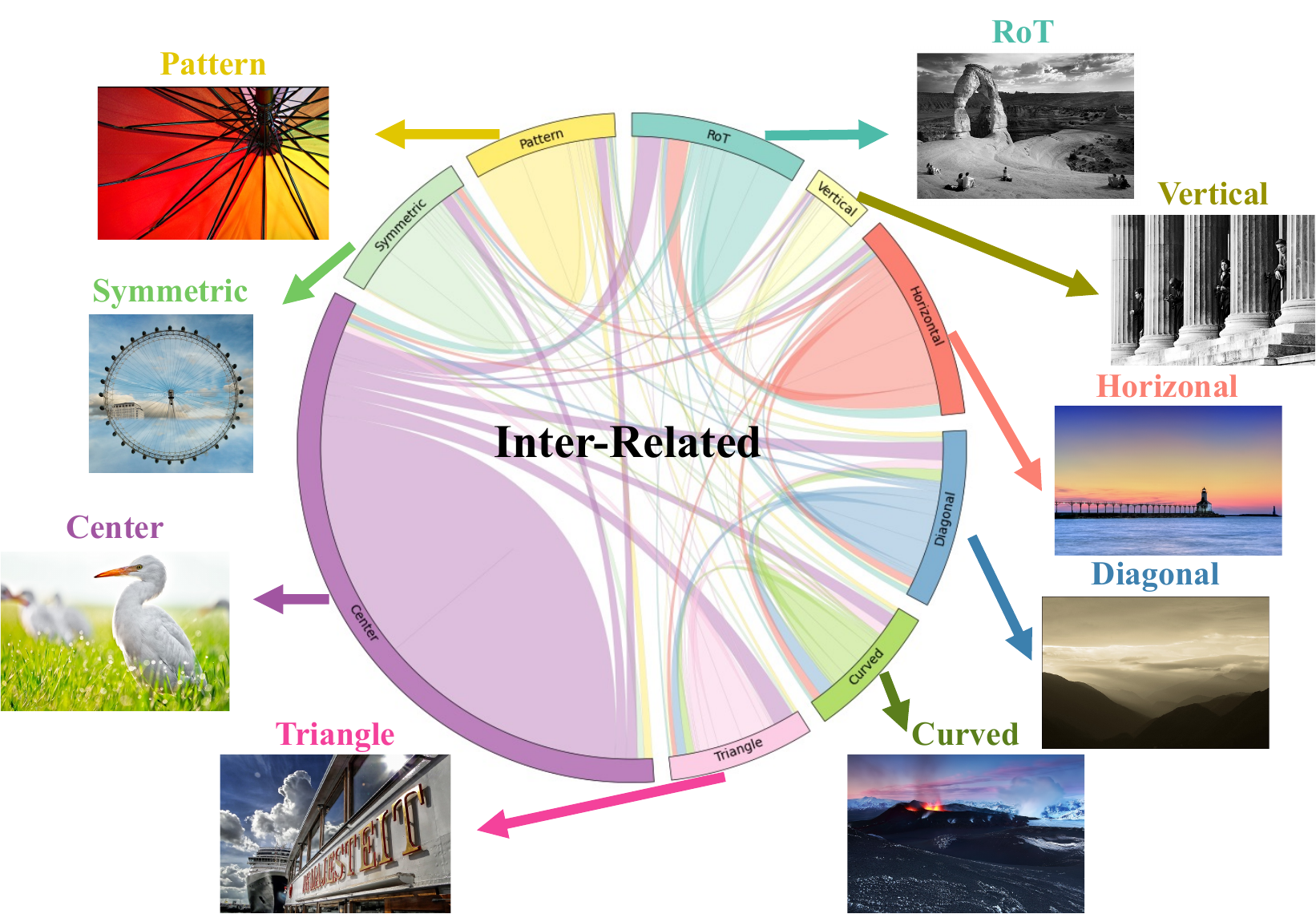}
\caption{{Distribution of compositional layouts across CAD.}}
\label{fig:figure_layout}
\vspace{-.1in}
\end{figure}

\begin{table}[!t]
	\caption{
 {Distribution of composition categories in CAD, where ``RoT" denotes Rule of Thirds.}
 }\label{tab:dist}
 \vspace{-.1in}
	\centering
		\resizebox{0.9\linewidth}{!}{
			\begin{NiceTabular}{l|cc|cc|c}
                      \CodeBefore
            \rowcolor{babyblue!30}{12}
            \Body
				\toprule
				\multirow{2}{*}{Composition}&\multicolumn{2}{c}{Original}&\multicolumn{2}{c}{Out-painted}&\multirow{2}{*}{Total}\\
				\cmidrule(lr){2-3}\cmidrule(lr){4-5}
				&AVA&UnSplash&AVA&UnSplash&\\
                \midrule
                RoT &5376&2203&15050 & 5652 & 20702 \\
                Vertical&1897&707& 6023 & 1990 & 8013 \\
                Horizontal&4322&5755& 19706 & 8428 & 28134 \\
                Diagonal&5339&2455& 15023 & 487 & 15510 \\
                Curved &2998&1330& 10661 & 4447 & 15108 \\
                Triangle &5147&2646& 11816 & 3172 & 14988 \\
                Center &19665&8066& 80150 & 13162 & 93312 \\
                Symmetric &1669&1360& 16105 & 5562 & 21667 \\
                Pattern &3182&449& 17588 & 2658 & 20246 \\
                \textbf{Total}&49595&24971& 192122 & 49942 & \textbf{242064} \\
                \bottomrule
		\end{NiceTabular}}
   \vspace{-.1in}
\end{table}

\begin{figure}[!t]
\centering
         \includegraphics[width=0.9\linewidth]{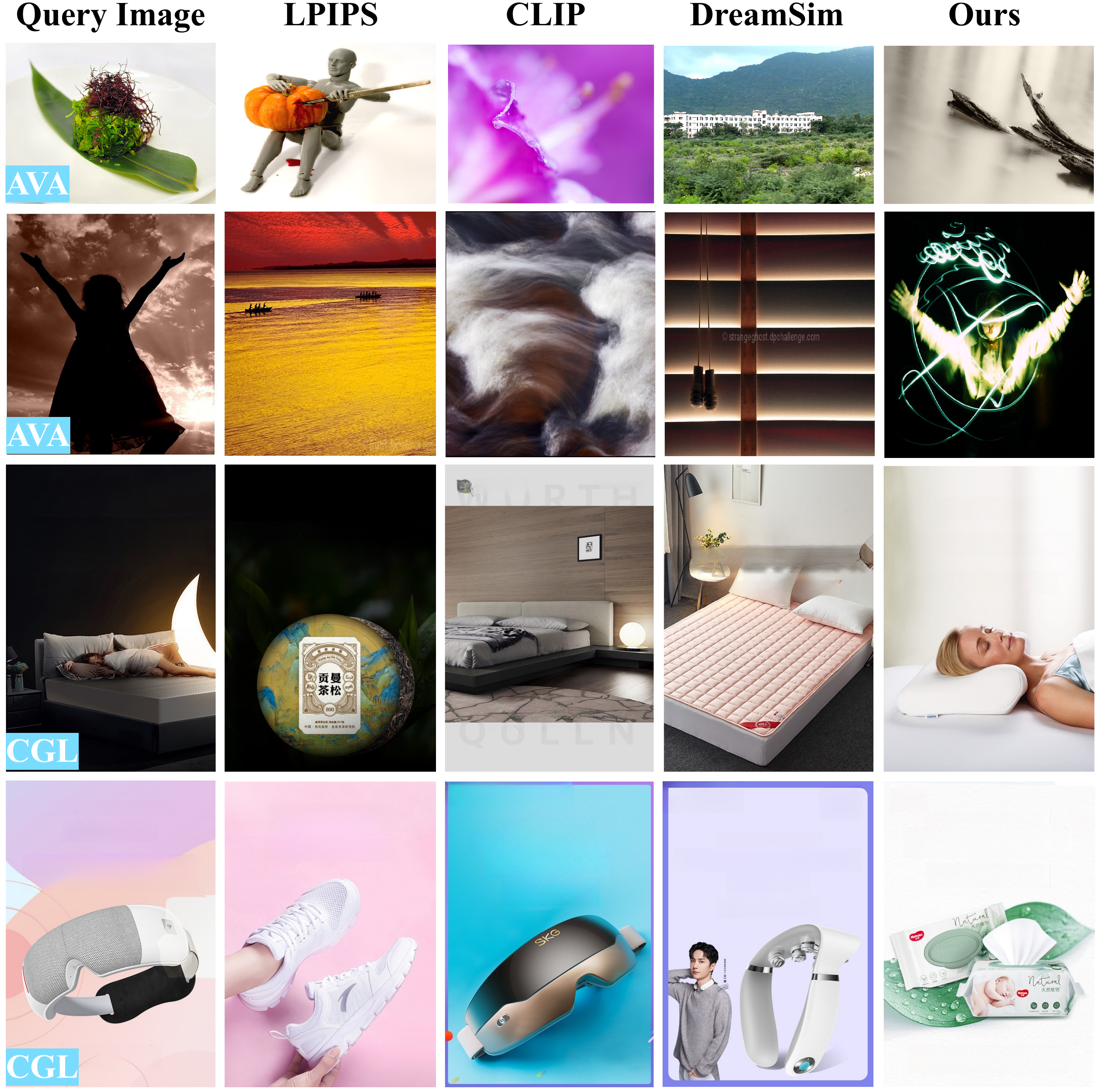}
\caption{
{Retrieval comparison on CGL and AVA datasets. Our approach exhibits superior image retrieval performance compared to other methods by prioritizing line composition, yielding matches with enhanced compositional relevance.}
}
\label{fig:re}
\vspace{-.1in}
\end{figure}


{To create our composition-aware dataset (CAD), we source professional images from AVA~\cite{6247954} and Unsplash Lite~\cite{unsplash2023}. From AVA, we select the top 55,000 images based on their aesthetic scores. The Unsplash Lite dataset contributes 25,000 high-quality, nature-themed photographs, which are available for both commercial and non-commercial use. Using these 80,000 curated professional images as a foundation, we generate 242,000 synthetic images that meet our quality standards through automatic filtering~\cite{hong2024learning}. The distribution of compositional layouts and categories in CAD is shown in~\cref{fig:figure_layout,,tab:dist}.}

\subsection{Implementation details}

{Following cDETR~\cite{meng2021conditional} and recent works~\cite{jia2022rethinking,hong2024learning}, we optimize our model using AdamW optimizer with a weight decay of $10^{-4}$. The learning rate is set to 
$10^{-4}$, with a reduced rate of 
$10^{-5}$ for the CNN backbone. The model trains for 500 epochs. In the weakly-supervised setting with our curated CAD, we divide training into two stages: Stage~1 (first 100 epochs) initializes the model weights, while Stage~2 (remaining 400 epochs) involves crop prediction and dynamic ranking to generate diverse pseudo-labels.}

{\textbf{Evaluation Metrics.}
We adopt three evaluation metrics, including Intersection-over-Union (IoU), boundary displacement (Disp), and top-N accuracy (ACC$_{K/N}$), following~\cite{hong2024learning, yang2023focusing, zhang2022human}. IoU and Disp provide objective and consistent comparisons, while ACC$_{K/N}$ reflects human perception. Specifically, for ACC$_{K/N}$, we define the best crops of an image as those ranked within the top-N by mean opinion scores (MOS) from human ratings. ACC$_{K/N}$ then measures how many of the top-K predicted crops fall within this top-N MOS set. This makes ACC$_{K/N}$ highly correlated with user study results.
Following ~\cite{su2024spatial, wang2023image}, we report the average top-$k$ accuracy ($\overline{\text{ACC}_{k}}$) for $k=5$ and $k=10$. When predicted views do not align exactly with predefined grid views, we consider two crops equivalent if their IoU exceeds a threshold of $\epsilon = 0.85$, as in ~\cite{liu2023beyond, jia2022rethinking}.
}

\subsection{Comparative assessment}


{We first conduct comparative analysis on retrieval approaches and then evaluate our ProCrop model performance in both supervised and weakly-supervised settings.}

{\textbf{Retrieval approaches analysis.} We compare our SAM-based retrieval against SOTA embeddings~(DreamSim \cite{fu2024dreamsim}, OpenCLIP~\cite{cherti2023reproducible}) and established learned metrics like LPIPS~\cite{zhang2018unreasonable}. Our evaluation uses examples from CGL~\cite{zhou2022composition} and AVA~\cite{6247954} datasets, where for each query image, we compute similarities across the dataset and retrieve the nearest neighbors based on each metric. As shown in~\cref{fig:re}, existing methods either focus on fine-grained visual features (LPIPS emphasizing background color) or broader semantic attributes (DreamSim and OpenCLIP focusing on object categories). In contrast, our SAM-based retrieval uniquely excels at identifying compositional similarities across diverse visual styles, demonstrating effective generalization without relying on category information.}


\begin{figure*}[!t]
\centering
         \includegraphics[width=0.9\textwidth]{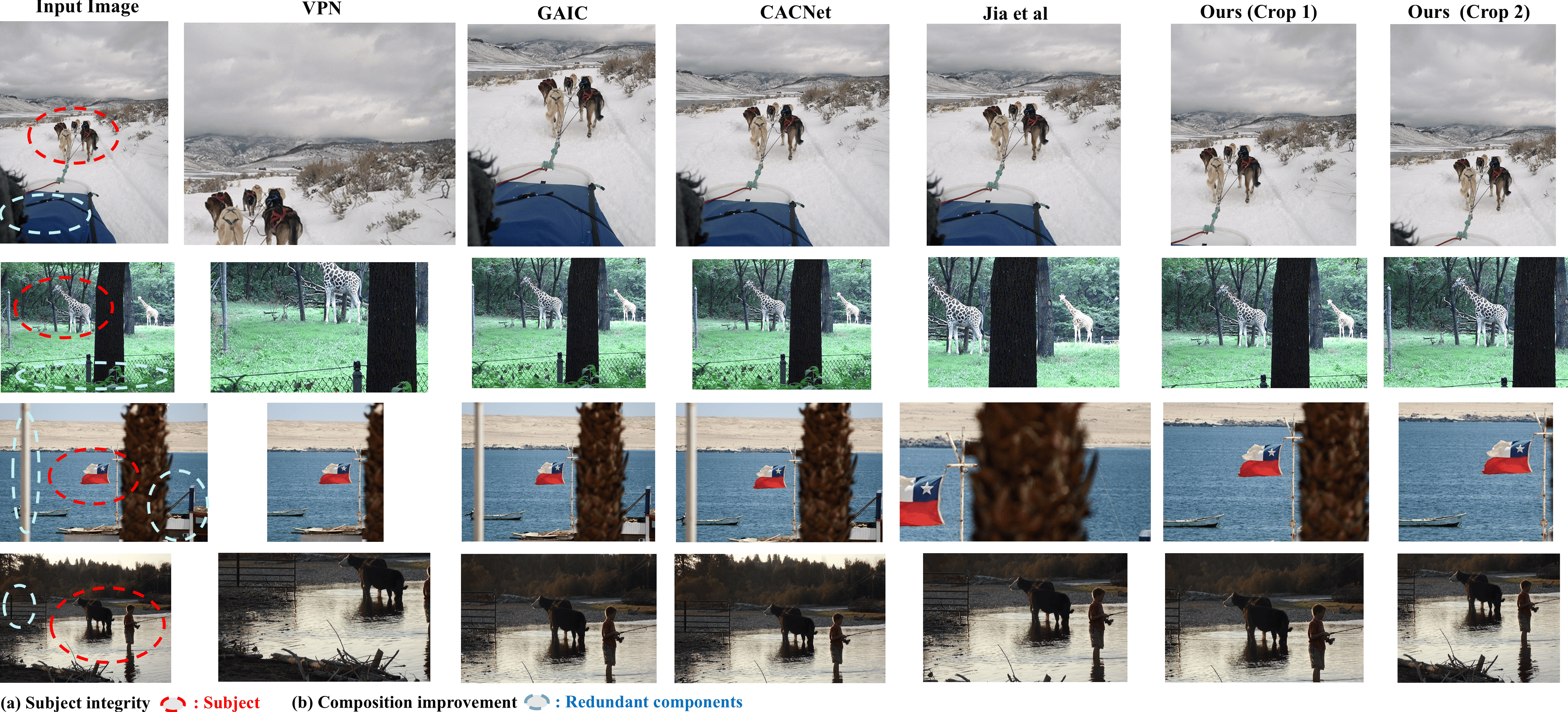}
\caption{
{Qualitative comparison of cropping results. Our approach preserves primary subjects (red boxes) while removing redundant elements (blue boxes), maintaining subject integrity and enhancing aesthetic composition.}}
\label{fig:figure5}
\vspace{-.1in}
\end{figure*}

\begin{table}[t]
	\caption{Comparison under supervised setting. We compute our metrics and report comparative results based on~\cite{liu2023beyond,jia2022rethinking,wang2023image,su2024spatial}.}.\label{tab:tab2}
 \vspace{-.1in}
	\centering
		\resizebox{0.9\linewidth}{!}{
			\begin{NiceTabular}{lcccc}
                    \CodeBefore
            \rowcolor{babypink!30}{13}
            \rowcolor{babypink!30}{25}
            \Body
                \toprule
				\multirow{2}{*}{Methods} & \multicolumn{4}{c}{GAICv2}\\
                \cmidrule(lr){2-5} 
                &ACC$_{1/5}$($\uparrow$) & $\overline{\text{ACC}}_{5}$($\uparrow$)&ACC$_{1/10}$($\uparrow$)&$\overline{\text{ACC}}_{10}$($\uparrow$)\\
                \midrule
                {A2-RL~\cite{li2018a2}}&23.2&26.4&{39.5}&40.1\\ 
                {VFN~\cite{chen2017learning}}&26.6&{26.4}&40.6&{40.1}\\ 
                {VEN~\cite{wei2018good}}&37.5&{50.5}&35.5&{48.6}\\ 
                {CGS~\cite{li2020composing}}&63.0&{59.7}&81.5&77.8\\ 
                {GAICv2~\cite{zeng2020grid}}&68.2&{63.1}&84.4&{81.6}\\ 
                {TransView~\cite{pan2021transview}}&69.0&{63.9}&85.4&{82.4}\\ 
                {HCIC}~\cite{zhang2022human}&-&63.8&-&81.3\\ 
                {Jia \textit{et al}~\cite{jia2022rethinking}}&\underline{85.0}&-&\underline{92.6}&-\\ 
              {Chao \textit{et al}~\cite{wang2023image}}&70.0&{\underline{64.8}}&86.8&{\underline{83.3}}\\ 
            {S$^2$CNet}~\cite{su2024spatial}&-&{64.0}&-&{82.7}\\ 
                Ours ($\epsilon=0.85$)&\textbf{85.4}&\textbf{81.8}&\textbf{94.2}&\textbf{91.2}\\
				\toprule
				\multirow{2}{*}{Methods}&\multicolumn{2}{c}{GAICv1} &\multicolumn{2}{c}{FLMS}\\
				\cmidrule(lr){2-3}\cmidrule(lr){4-5} 
				&ACC$_{1/5}$($\uparrow$) & \multicolumn{1}{c}{ACC$_{1/10}$($\uparrow$)} & IOU ($\uparrow$)&Disp($\downarrow$)\\
                \midrule
                {A2-RL~\cite{li2018a2}}&23.0&{38.5}&0.821&0.045\\
                {VFN~\cite{chen2017learning}}&27.0&{39.0}&0.577&0.124\\ 
                {VPN~\cite{wei2018good}}&40.0&{49.5}&0.835&-\\ 
                {VEN~\cite{wei2018good}}&40.5&{54.0}&0.837&0.041\\ 
                {CGS~\cite{li2020composing}}&63.0&{81.5}&0.836&0.039\\ 
                {GAICv1~\cite{zeng2019reliable}}&53.5&{71.5}&-&-\\ 
                {ASM-Net~\cite{tu2020image}}&54.3&{71.5}&-&-\\ 
                {Jia \textit{et al}~\cite{jia2022rethinking}}&\underline{81.5}&{\underline{91.0}}&0.838&\underline{0.037}\\ {UNIC~\cite{liu2023beyond}}&-&{-}&\underline{0.840}&\underline{0.037}\\ 
                Ours($\epsilon=0.85$)&\textbf{86.0}&\textbf{94.5}&\textbf{0.843}&\textbf{0.036}\\
                \bottomrule
		\end{NiceTabular}}
  \vspace{-.1in}
\end{table}

{\textbf{Evaluation under supervised setting.} We evaluate our model against various baselines trained on GAICv1~\cite{zeng2019reliable}, GAICv2~\cite{zeng2019reliable}, and CPC~\cite{wei2018good}. For models trained on GAICv1 and GAICv2, we evaluate using $\text{ACC}_{\text{1/5}}$ and $\text{ACC}_{\text{1/10}}$ on their respective test sets. For models trained on CPC, we measure IoU on the FLMS dataset. To ensure fair comparison, we exclude text embeddings from feature fusion. A key feature of our approach is the integration of the retrieval module, which fetches 10 similar images from the top-rated 55,000 images in AVA during both training and inference. \cref{tab:tab2} shows that our method significantly outperforms previous approaches across all datasets and metrics, demonstrating the effectiveness of guidance from retrieved professional image compositions.}

\begin{table}[!t]
	\caption{
     {Comparison with supervised and weakly-supervised (WS) benchmarks on SACD dataset. The comparative results are borrowed from~\cite{hong2024learning}. N denotes the number of crop proposals.}
    }\label{tab:tab4}
 \vspace{-.1in}
	\centering
		\resizebox{0.9\linewidth}{!}{
			\begin{NiceTabular}{lcccc}
            \CodeBefore
            \rowcolor{babyblue!20}{13}
            \rowcolor{babyblue!20}{14}
            \rowcolor{babypink!30}{12}
            \Body
				\toprule 
				Methods & Trained on & WS &IOU & Disp \\
                \midrule
                LVRN~\cite{lu2019listwise}&CPC&\ding{55}&{0.6962}&0.0765\\
                GAIC~\cite{zeng2020grid}&GAICD&\ding{55}&0.7124&0.0696\\
                CACNet~\cite{li2020composing}&FCDB,KUPCP&\ding{55}&0.7109&0.0716\\
                HCIC~\cite{zhang2022human}&GAICD&\ding{55}&0.7120&0.0683\\
               HCIC~\cite{zhang2022human}&CPC&\ding{55}&0.7109&0.0712\\
                
                {VPN~\cite{wei2018good}}&CPC+AADB&\ding{55}&0.7164&0.0663\\
                {VPN~\cite{wei2018good}}&Flickr&$\checkmark$&0.6690&0.0887\\
                {VPN~\cite{wei2018good}}&Unsplash&$\checkmark$&0.6555&0.0775\\
                {Gencrop~\cite{hong2024learning}}&Unspash&$\checkmark$&0.7301&0.0632\\
                \midrule
                {Ours (w/o rtr.)}&CAD&$\checkmark$&0.7035&0.0722\\
                {Ours (N=1)}&CAD&$\checkmark$&\textbf{0.7303}&\textbf{0.0610}\\
                {Ours (N=2)}&CAD&$\checkmark$&0.7546&0.0541\\
                {Ours (N=3)}&CAD&$\checkmark$&0.7678&0.0506\\
				\bottomrule
		\end{NiceTabular}}
  \vspace{-.1in}
\end{table}



{\textbf{Evaluation under weakly-supervised (WS) setting.}
We evaluate ProCrop, trained on our large-scale  CAD dataset, on the unseen subject-aware SACD dataset (zero-shot transfer). \cref{tab:tab4} compares our approach with previous subject-aware methods on supervised and WS benchmarks. Unlike prior methods using sliding-window ensembles that are later combined into a single output, ProCrop generates diverse, aesthetic crops in a single pass.
With 90 predicted crops, our highest-scoring crop outperforms ensemble outputs of existing methods (e.g., GAIC, CACNet, Gencrop) in both IOU and Disp metrics. Our method further excels in generating multiple effective crop candidates. Notably, our full model with the retrieval module significantly outperforms the variant without retrieval, highlighting the effectiveness of retrieval guidance in this WS scenario.}

{We visually compare crops produced by our method with those generated by existing approaches. To evaluate crop quality objectively, we adopt two criteria: adherence to the subject integrity principle~\cite{jia2022rethinking, hong2021composing, hong2024learning}, which requires preserving the main subject (e.g., a person) naturally in the cropped result, and enhancement of aesthetic composition by eliminating redundant elements to achieve a more visually appealing result. \cref{fig:figure5} illustrates these comparisons. Notably, our predicted crops effectively capture the salient subject while significantly enhancing the overall aesthetic quality of the image.}


\begin{figure}[!t]
     \centering
     \begin{subfigure}[b]{0.47\linewidth}
         \includegraphics[width=\linewidth]{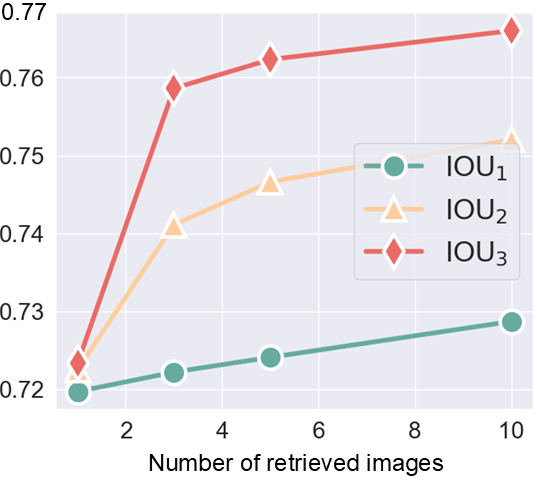}
         \caption{}
        \label{fig:bboxshift_a}
     \end{subfigure}
     \begin{subfigure}[b]{0.515\linewidth}
\includegraphics[width=\linewidth]{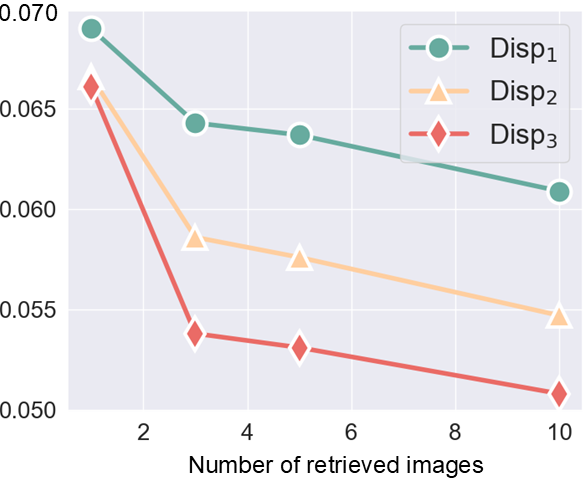}
         \caption{}
     \end{subfigure}
     \vspace{-.2in}
\caption{
{Impact of retrieval count
. We show the relationship between IOU and Disp versus the number of retrieved images. IOU$_{i}$/Disp$_{i}$ denotes evaluation on the top $i$ crop proposals.}
}
\label{fig:retrieve_number}
\vspace{-.1in}
\end{figure}

\begin{table}[!t]
	\caption{
    {ProCrop performance across different retrieval sources. Baseline results correspond to Jia \textit{et al.}~\cite{jia2022rethinking}.}
    }\label{tab:tab3}
    \vspace{-.1in}
	\centering
		\resizebox{0.9\linewidth}{!}{
			\begin{NiceTabular}{cccc|cc}
            \CodeBefore
\rowcolor{babypink!30}{8}
\rowcolor{babyblue!20}{4}
\rowcolor{babyblue!20}{5}
\rowcolor{babyblue!20}{6}
\rowcolor{babyblue!20}{7}
\Body
				\toprule
				\multicolumn{4}{c}{Retrieval}&\multicolumn{2}{c}{GAICv1}\\
				\cmidrule(lr){1-4}\cmidrule(lr){5-6}
				Retrieve set&Set size&Image&Annotation&\multicolumn{1}{c}{ACC$_{5}$}&\multicolumn{1}{c}{ACC$_{10}$}\\
                \midrule
                -&-&-&-&{0.815}&0.910\\
                GAICv1&1000&$\checkmark$&\ding{55}&{0.805}&0.920\\
                GAICv1&1000&\ding{55}&$\checkmark$&{0.820}&0.920\\
                GAICv1&1000&$\checkmark$&$\checkmark$&{0.834}&0.915\\
                CPC&10000&$\checkmark$&$\checkmark$&{\underline{0.840}}&\underline{0.940}\\
                AVA&55000&$\checkmark$&\ding{55}&{\textbf{0.860}}&\textbf{0.945}\\
                \bottomrule
		\end{NiceTabular}}
        \vspace{-.1in}
\end{table}

\subsection{Ablation study}

We present ablations on retrieval sources, number of retrieved images, and our model components. Further ablations on the retrieval encoder, feature alignment, crop number, efficiency, transferability, and retrieval-prediction relationship are analyzed in Appendix A.

{\textbf{Retrieval from different datasets.} 
\cref{tab:tab3} presents five ablation studies comparing our method with the second-best approach by Jia \textit{et al.}~\cite{jia2022rethinking}, with all models trained on the GAICv1 dataset. When retrieving only GAICv1 images, our performance was comparable to the baseline, likely due to non-professional retrieved images. However, utilizing GAICv1 encodede image-label pairs improved performance beyond the baseline. Expanding the retrieval set to include CPC images led to further improvements (ACC$_5$: 0.840, ACC$_{10}$: 0.940), benefiting from diverse compositional elements. Finally, incorporating the professional AVA dataset, even without label pairs, achieved the highest performance  (ACC$_5$: 0.860, ACC$_{10}$: 0.945). These results underscore the importance of diverse and high-quality retrieval sources in enhancing the performance of our ProCrop method.}


{\textbf{Impact of retrieval image count.} \cref{fig:retrieve_number} illustrates how the number of retrieved images affects model performance. Models trained on our CAD dataset and evaluated on the SACD dataset show similar values for IOU$_1$, IOU$_2$, and IOU$_3$, when only one image is retrieved. As the retrieval count increases, greater diversity in crop compositions leads to significant improvements in both IOU and Disp metrics. Performance stabilizes around ten retrieved images, benefiting from diverse layout information.}


\begin{table}[!t]
	\caption{
    Ablation studies of ProCrop components under weakly-supervised setting. N denotes the number of crop proposals. The results are reported on the SACD dataset. 
    }\label{tab:tab6}
    \vspace{-.1in}
	\centering
		\resizebox{0.9\linewidth}{!}{
			\begin{NiceTabular}{c|c|c|cccc}
          \CodeBefore
\rowcolor{babypink!30}{4}
\rowcolor{babypink!30}{7}
\Body
				\toprule
				{Retrieve}&{Text}&Metric&{N=1}&N=2&N=3&Avg\\
                \midrule
{\ding{55}}&{\ding{55}}&\multirow{3}{*}{IOU ($\uparrow$)}&{0.7035} & 0.7114 & 0.7160 & 0.7103\\
{$\checkmark$}&{\ding{55}}&&{\underline{0.7287}} & \underline{0.7520} & \underline{0.7660} & \underline{0.7489}\\
{$\checkmark$}&{$\checkmark$}& &{\textbf{0.7303}} & \textbf{0.7546} & \textbf{0.7678} & \textbf{0.7509}\\
\midrule
{\ding{55}}&{\ding{55}}&\multirow{3}{*}{Disp ($\downarrow$)}&{0.0722}& 0.0647 & 0.0632 & 0.0667\\
{$\checkmark$}&{\ding{55}}&&{\textbf{0.0609}}& \underline{0.0547} & \underline{0.0508} & \underline{0.0555}\\
{$\checkmark$}&{$\checkmark$}&&{\underline{0.0610}} & \textbf{0.0541} & \textbf{0.0506} & \textbf{0.0552} \\
				\bottomrule
		\end{NiceTabular}}
        \vspace{-.1in}
\end{table}

{\textbf{Components of ProCrop.}
\cref{tab:tab6} evaluates the effectiveness of ProCrop components in the weakly-supervised setting, focusing on image retrieval and text embeddings. Results show that incorporating image retrieval leads to notable improvements in average IoU (0.7489 vs. 0.7103) and Disp (0.0555 vs. 0.0667) metrics. The addition of text embeddings further enhances performance, demonstrating the effectiveness of our proposed strategies.}
\section{Conclusion}
{This work presents a novel composition-aware cropping framework that leverages professional images with similar aesthetic compositions. Our key contributions include a retrieval-based approach integrating features from professional images with query image embeddings, along with a large-scale compositional-aware cropping dataset. Through comprehensive evaluation across image retrieval, supervised, and weakly-supervised image cropping tasks, our results demonstrate state-of-the-art performance, showcasing robust and general applicability across various benchmarks compared to existing approaches.}


{
    \small
\bibliographystyle{ieeenat_fullname}
    \bibliography{main}
}


\twocolumn[
\centering
\Large
\textbf{ProCrop: Learning Aesthetic Image Cropping from Professional Compositions} \\
\vspace{0.5em}Supplementary Material \\
\vspace{1.0em}
]
In this supplementary material, we firstly present additional experiments and details to complement our main paper (\textbf{Ablations:} \Cref{sec:retrieval}). We conduct extensive ablation studies to analyze the impact of key components, including the retrieval encoder, feature alignment module, number of generated crops, time efficiency, computational cost, inference-time transferability, and the influence of retrieved images.
Next, we provide a detailed description of our datasets (\textbf{Datasets:} \Cref{sec:data}) , covering both our newly developed out-painted dataset and the diversity of the retrieved dataset.
We then elaborate on the text embedding (\textbf{Text:} \Cref{sec:text}), explaining the feature extraction procedure and the underlying rationale. We also evaluate the extracted layout features (\textbf{Layout:} \Cref{sec:layout}), provide visualizations, and compare them with existing saliency-based methods.
Finally, we discuss the limitations of our proposed model and explore its potential applications (\textbf{Discussion:} \Cref{sec:discuss}).

\appendix

\begin{table}[!h]
	\caption{
    \textbf{Encoder of retrieved images}: The "Memory" column indicates the memory consumption ratio of the cDETR encoder compared to the SAM encoder for processing retrieved images.
    }\label{tab:encoder}
    \centering
\resizebox{\linewidth}{!}{
\begin{NiceTabular}{cc|ccc}
\toprule
\multicolumn{2}{c}{Encoder of retrieved images.}&\multicolumn{3}{c}{GAICv2}\\
\cmidrule(lr){1-2}\cmidrule(lr){3-5}
cDETR&SAM&ACC$_{5}$&ACC$_{10}$&Memory\\
\midrule
\checkmark&\ding{55}&85.0&93.8&1.53\\
\ding{55}&\checkmark&85.4&94.2&1.00\\
\bottomrule
\end{NiceTabular}}
\end{table}

\begin{table}[!h]
	\caption{
   \textbf{Comparison of retrieval time}. AVA$_F$ denotes the full set of AVA dataset. 
    }\label{tab:time}
    \centering
\resizebox{\linewidth}{!}{
\begin{NiceTabular}{ccc|c}
\toprule
Dataset&Retrieve number&Size of database&Retrieve time\\
\midrule
GAICv2&10&2,626&0.094s\\
CPC&10&10,000&0.099s\\
AVA$_F$&10&255,000&0.954s\\
\bottomrule
\end{NiceTabular}}
\end{table}

\section{Additional ablations} \label{sec:retrieval}
In this section, we present additional ablation studies on retrieval details to evaluate the impact of the retrieval encoder, assess time efficiency, and examine flexibility during test-time inference. Finally, we provide more examples to show the connection between retrieved images and crop proposals predicted by our model.

\subsection{Encoder of retrieved images} 
We evaluate the model's performance by using different encoders for retrieving images. 
Specifically, we compare the cDETR encoder, used for processing query images, with the SAM encoder utilized in our ProCrop framework.
Our model is trained and evaluated on the GAICv2 dataset, using the professional subset of AVA as the retrieval set.
As shown in \Cref{tab:encoder}, the SAM encoder achieves slightly better performance while requiring less memory.
This improvement can be attributed to SAM's extensive pretraining on a large dataset, which equips it with a robust ability to extract boundary features across various unlabeled images. 
For efficiency and effectiveness, we adopt SAM as the encoder for retrieved images in our approach.

\subsection{Feature alignment}
We elaborate on the details of feature alignment and conduct ablation study on this module.

\noindent\textbf{Description:} We follow Daichi \textit{et al}.~\cite{horita2024retrieval} for feature alignment.
We denote this query image feature as $\bar{f}_I\in\mathbb{R}^{p\times d}$, where $p$ represents the flattened spatial dimension specific to this encoder. To effectively fuse $\bar f_I$ with $\bm R$, we employ a learnable projection head $\Pi(\cdot)$ that transforms $\bm R$ to match the spatial-channel dimensions of $\bar f_I$. The final feature fusion is achieved through:
\begin{equation}
f_R = \text{Concat}(\bar f_I, \Pi(\bm R), f_c),
\end{equation}
where $f_c$ denotes the cross-attended feature obtained by using $\bar f_I$ as the query and $\bm R$ as both key and value. This design enhances the interaction between the input canvas and reference layouts.

\noindent\textbf{Ablation on feature alignment}
We conduct three groups of experiments on the SACD dataset. The first implementation (\#1) does not use any retrieved features, i.e., $f_R = f_I$. The second implementation (\#2, "concat") directly concatenates the features of the query image and the retrieved images without incorporating cross-attended features, i.e., $f_R = \text{Concat}(\bar f_I, \Pi(\bm R))$. The third implementation (\#3, "concat + CA") further integrates cross-attended features into the concatenation, with $f_R = \text{Concat}(\bar f_I, \Pi(\bm R), f_c)$.  

As shown in \Cref{tab:align}, directly concatenating retrieved features (\#2) improves performance over the baseline (\#1). Incorporating cross-attended features in \#3 leads to further performance gains, demonstrating the benefit of enhanced interaction between the retrieved and query features.  
\begin{table}[!t]
	
    \centering
    \caption{
   \textbf{Ablations of our feature alignment methods}. "concat" (concatenate) refers to directly concatenating the features of the retrieved and query images, while "CA" (cross-attention) further employs cross-attended feature for fusion.
    }\label{tab:align}
		\resizebox{0.8\linewidth}{!}{
			\begin{NiceTabular}{c|cccc}
				\toprule
				Implementation & Concat&CA & Dice & Disp\\
                \midrule
                \# 1 & $\times$&$\times$&0.7035 &0.0722\\
                \# 2 & $\checkmark$ &$\times$&\underline{0.7203}&\underline{0.0631}\\ \# 3 & $\checkmark$&$\checkmark$     &\textbf{0.7287}&\textbf{0.0609}\\
				\bottomrule
		\end{NiceTabular}}
\end{table}


\subsection{The number of generated crops}
Theoretically, the number of anchors should exceed the maximum number of good crops across all images. Based on the ablation study results from Jia \textit{et al.}~\cite{jia2022rethinking}, we adopt a generation number of 90. Firstly, using very few anchors can be detrimental, likely because a small anchor set may not provide enough information for effectively learning good crops. Secondly, an excessively large number of anchors has only a minor negative impact. 
We refer readers to the supplementary material of ~\cite{jia2022rethinking} for more details.

\subsection{Time efficiency}
We compare retrieval times to evaluate the efficiency of retrieving images from databases of varying sizes. \Cref{tab:time} summarize our retrieved times. For this evaluation, we use GAICv2, CPC, and AVA as examples of small, medium, and large databases, respectively. From each database, we retrieve the 10 images with the most similar line compositions. As shown, leveraging the elastic search implementation by Hugging Face~\cite{ElasticSearch_Huggingface}, our retrieval process remains efficient across databases of all sizes. Even when retrieving from a database containing 255,000 images, our model achieves an acceptable retrieval time of 0.954 seconds.

\subsection{Computational cost}
We analyze the memory cost (\Cref{tab:encoder}) and the training time cost (\Cref{tab:time}). The training time is closely related to the size of the retrieval set. As shown in \Cref{tab:time}, the retrieval time for 10 images ranges from 0.1 to 1 second, depending on the retrieval set size, which varies from 10,000 to 255,000. For larger retrieval sets, we can pre-compute the retrieval relationships to save time. Since our approach only involves inference rather than training the SAM encoder, it introduces a manageable memory cost, which is even significantly lower than that of cDETR.
\begin{table}[!t]
	\caption{
   \textbf{Inference-time transfer of retrieve sets:} AVA$_P$ denotes the subset of AVA professional images.
    }\label{tab:time}
    \centering
\resizebox{\linewidth}{!}{
\begin{NiceTabular}{cccc|cc}
\toprule
\multicolumn{4}{c}{Retrieve set}&\multicolumn{2}{c}{GAICv2}\\
\cmidrule(lr){1-4}\cmidrule(lr){5-6}
Train&Test&Size&Professional&ACC$_5$&ACC$_{10}$\\
\midrule
\multirow{3}{*}{AVA$_P$}
&\multicolumn{1}{|c}{CPC}&10,000&\ding{55}&85.2&93.2\\
&\multicolumn{1}{|c}{UnSplash-lite}&25,000&$\checkmark$&85.8&93.6\\
&\multicolumn{1}{|c}{AVA$_P$}&55,000&$\checkmark$&85.4&94.2\\
\bottomrule
\end{NiceTabular}}
\end{table}
\begin{figure}[!t]
\centering
         \includegraphics[width=\linewidth]{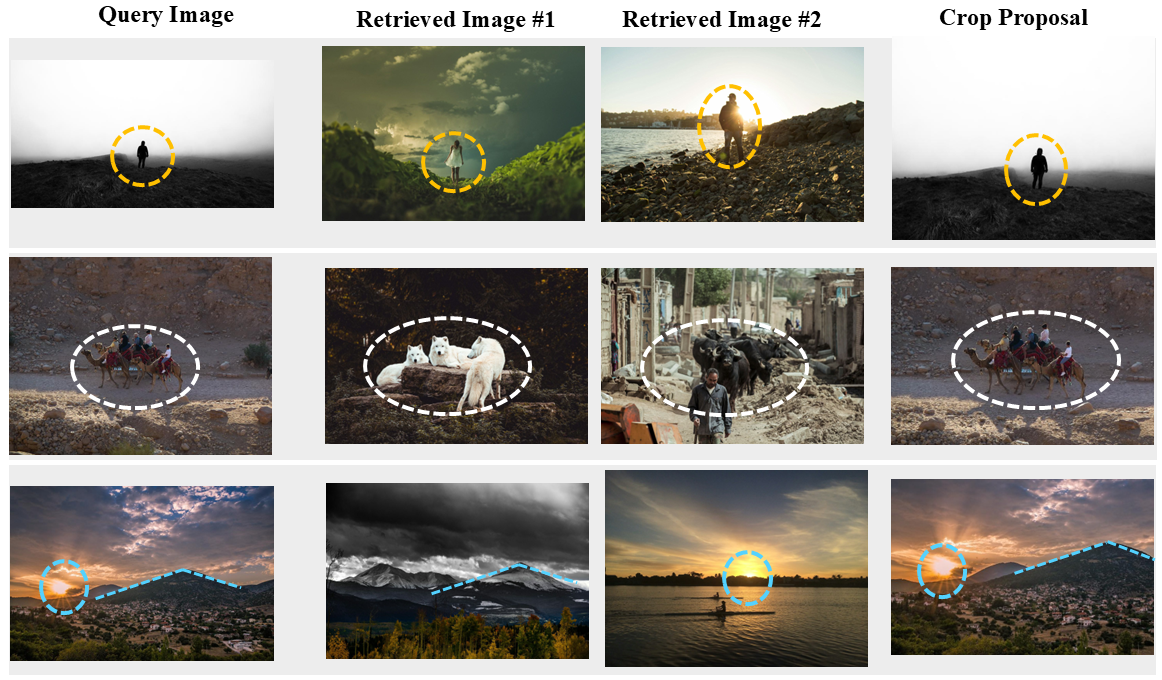}
\caption{
\textbf{Relationship between retrieved images and crop proposals.}  Unsplash-lite dataset is taken as the retrieve set for illustration.}
\label{fig:retrieve}
\end{figure}
\begin{figure*}[!t]
\centering
        \includegraphics[width=\textwidth]{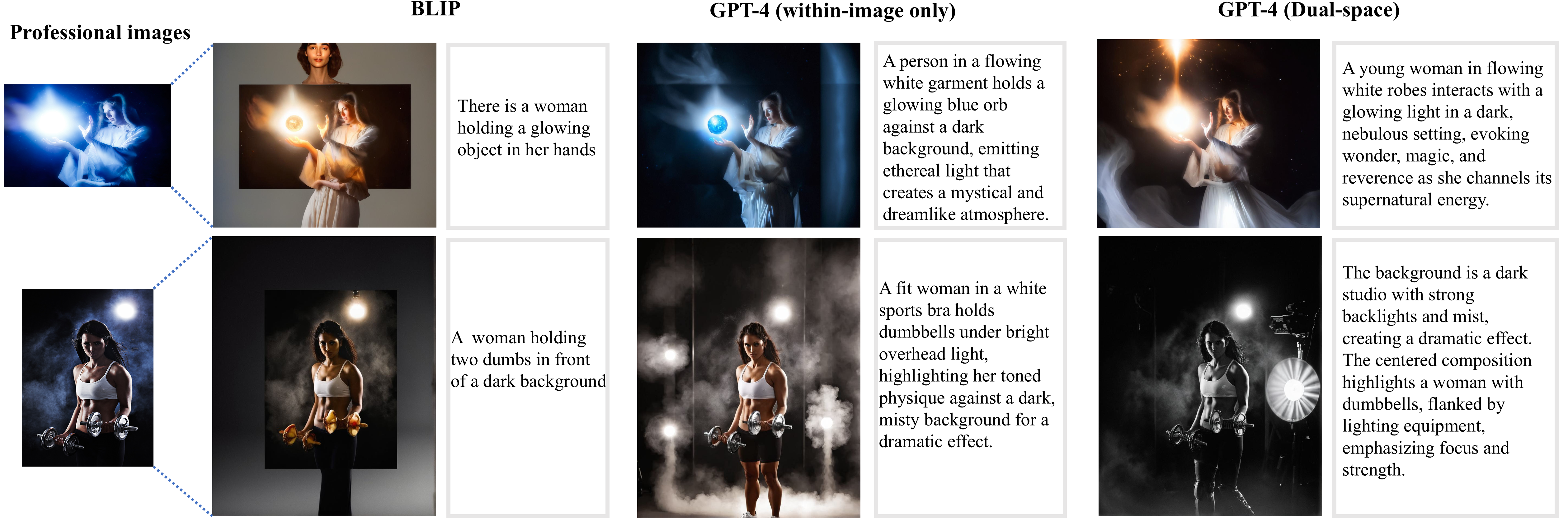}
\caption{
{\textbf{Illustration of text descriptions and corresponding outpainted results.} The text descriptions are generated using BLIP, GPT-4 (within-image only), and GPT-4 (dual-space understanding), respectively.}}
\label{fig:text}
\end{figure*}
\begin{figure}[!t]
\centering
        \includegraphics[width=\linewidth]{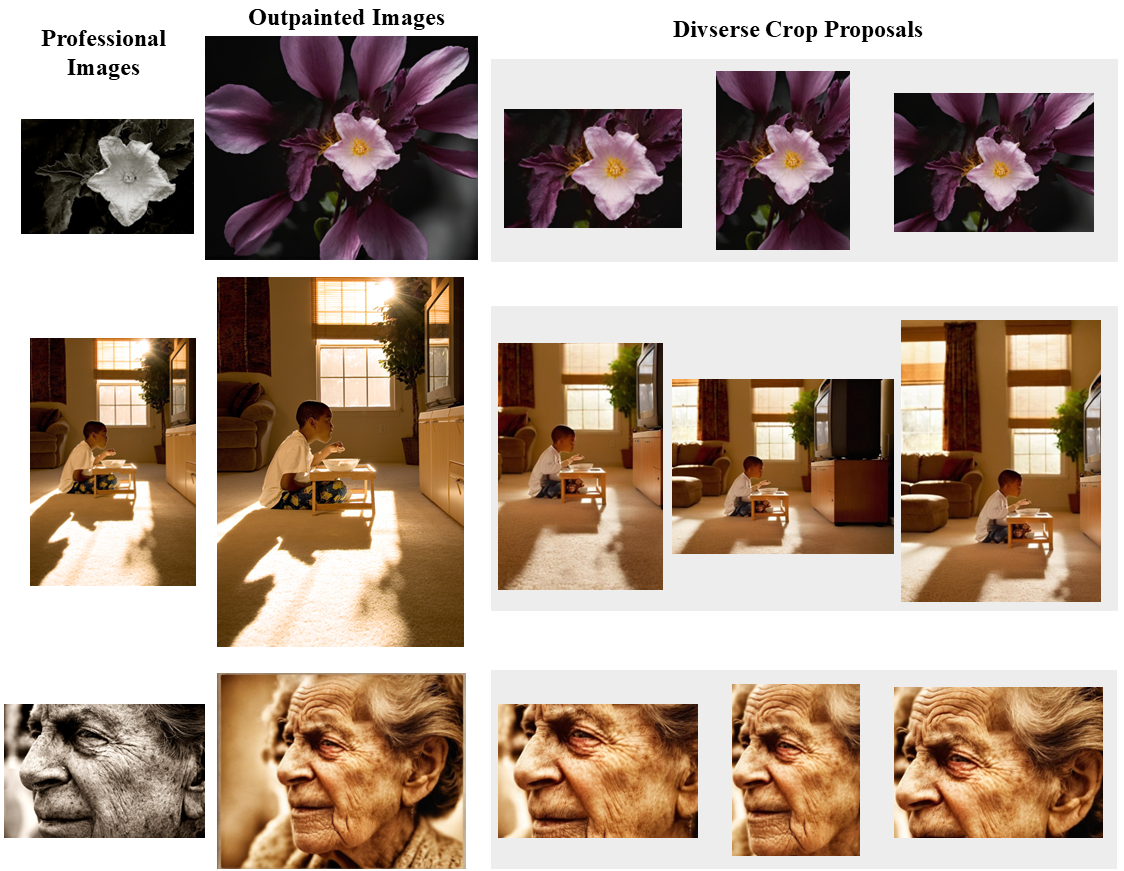}
\caption{
{\textbf{Illustration of more out-painted examples}: the visualization of out-painted images and their diverse crop proposals.}}
\label{fig:outpaint}
\end{figure}
\subsection{Inference-time transfer of retrieval sets}
\Cref{tab:time} illustrates the impact of changing the retrieval database during inference. The model is trained on the GAICv2 dataset using AVA$_P$ (the professional subset ranked by aesthetic scores for the top 55,000 images) and tested on the GAICv2 test set. During inference, the retrieval datasets include CPC, Unsplash-lite, and AVA$_P$. 

When CPC is used as the retrieval set, the model's performance drops on both the ACC$_5$ and ACC$_{10}$ metrics, likely due to the less professional and aesthetically pleasing nature of the CPC image compositions. In contrast, using professional photograph database Unsplash-lite as the retrieval set achieves performance comparable to AVA$_P$, demonstrating the model's transferability. This suggests the method's adaptability, as the retrieval dataset can be changed during inference without the need for retraining the model.

\subsection{Influence of retrieved images}
\textbf{Connection with predicted crops:} \Cref{fig:retrieve} shows the query image, retrieved images, and the crop proposals generated by our model for the GAICv2 dataset, using Unsplash-lite as the retrieval set for demonstration. Although the retrieved images may not have an identical composition to the query image, they often share similar line compositions. By leveraging features from multiple retrieved images, our model effectively predicts reasonable crop proposals based on these references.

\noindent\textbf{Clarification:} We understand that aesthetic quality can be influenced by factors beyond just layout. We clarify that our use of retrieved layouts is intended to provide complementary information for cropping, rather than to fully determine aesthetic quality.

\section{Datasets}\label{sec:data}
\subsection{Developed dataset}
We detail the text generation process and present additional visual examples of our outpainted images along with their crop proposals.
For text generation, \Cref{tab:prompt} presents the prompts used to generate within-image descriptions and dual-space understanding of image descriptions. \Cref{fig:text} compares the text generated by BLIP, GPT-4 (within-image only), and GPT-4 (dual-space), along with the corresponding outpainting results for each. As show in \Cref{fig:text}, the outpainting results generated using the dual-space understanding descriptions are more realistic and contain more detailed features.

\begin{table}[!t]
	\caption{
   \textbf{Examples of GPT-4 prompts} for with-in image only and dual-space understanding text generation. 
    }\label{tab:prompt}
    \centering
\resizebox{\linewidth}{!}{
\begin{NiceTabular}{p{2cm}|p{7cm}}
\toprule
\multicolumn{1}{c}{Type}&\multicolumn{1}{c}{GPT prompt}\\
\midrule
Within-image only&What's in this image? Describe composition clearly. For example, point out the location, shape, size of objects within image in detail. Summarize it within 30 words.\\
\midrule
Dual-space understanding&Describe the background of image, and guess the composition out of input image. Then, describe the layout of whole image. Make the picture natural. Summarize it within 30 words.\\
\bottomrule
\end{NiceTabular}}
\end{table}
\Cref{fig:outpaint} presents additional examples from our CAD dataset, showcasing outpainted images alongside their diverse crop proposals, highlighting the versatility of our approach.

\begin{figure}[!t]
  \centering
  \begin{subfigure}[b]{0.44\linewidth}
         \includegraphics[width=\linewidth]{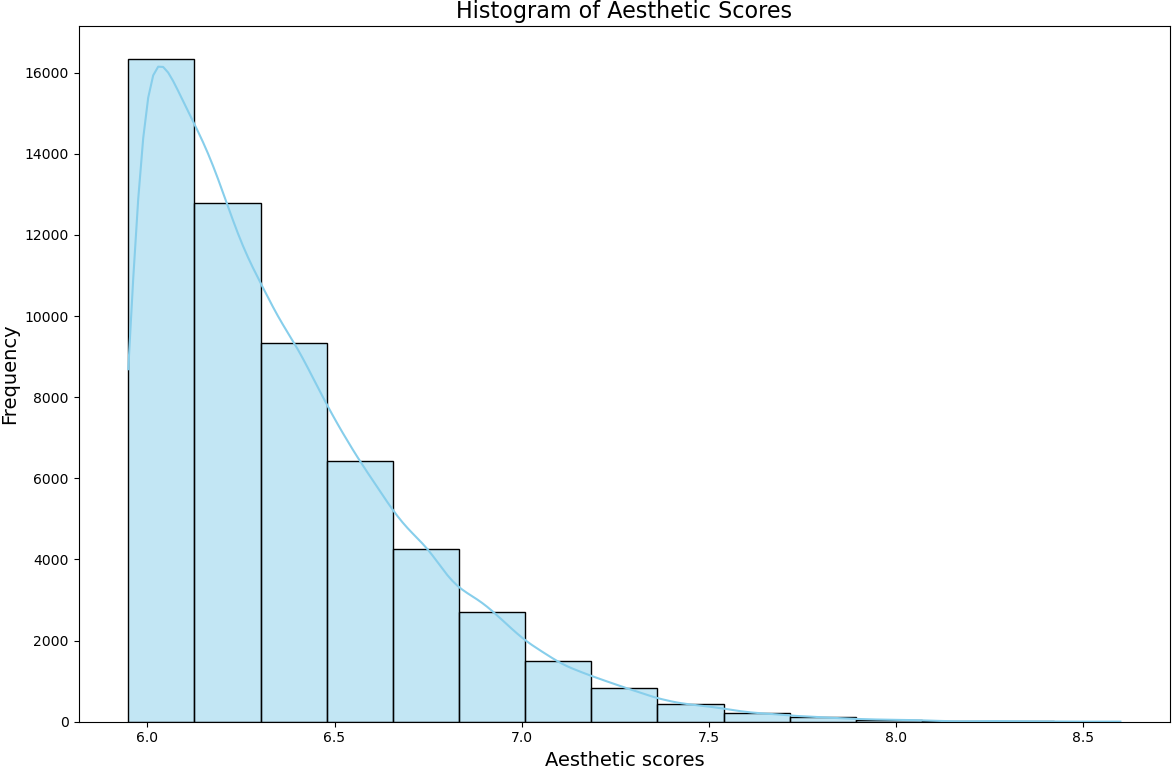}
         \caption{Aesthetic scores}
    \label{fig:short-b}
  \end{subfigure}
    \begin{subfigure}[b]{0.5\linewidth}
         \includegraphics[width=\linewidth]{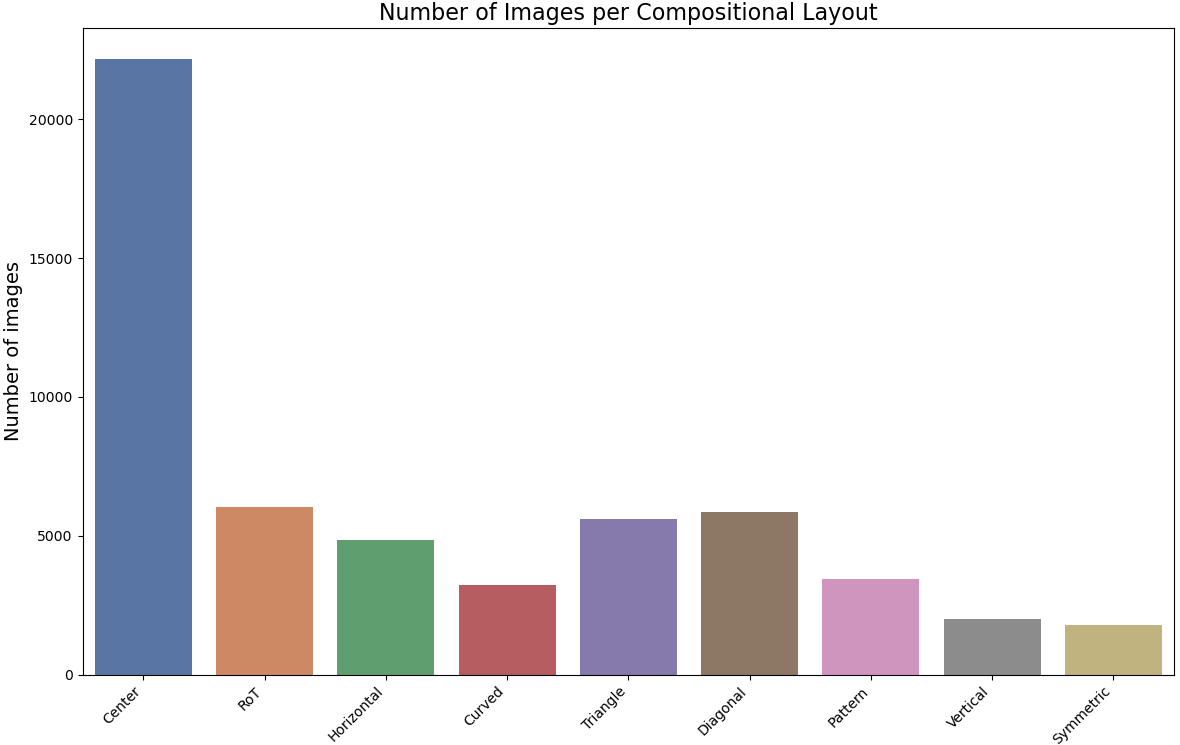}
         \caption{Layouts}
    \label{fig:short-d}
  \end{subfigure}
    \begin{subfigure}[b]{0.45\linewidth}
         \includegraphics[width=\linewidth]
         {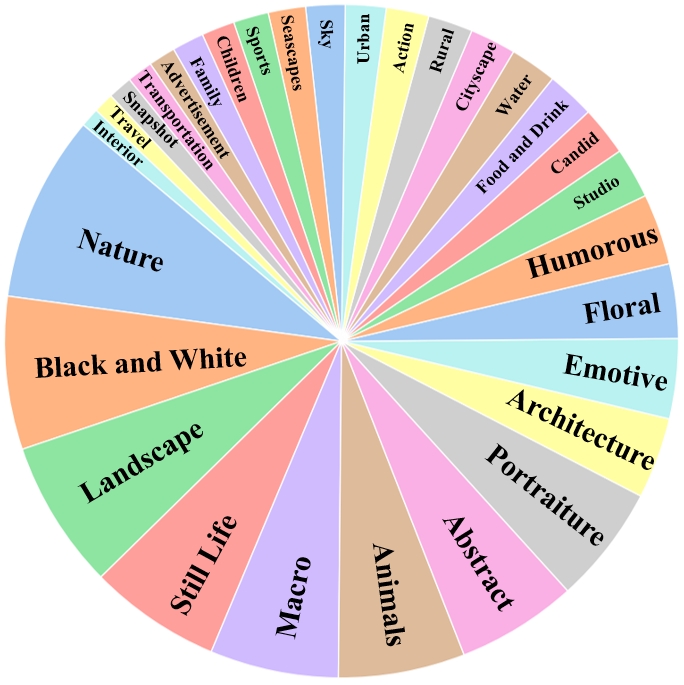}
         \caption{Semantic tags}
    \label{fig:short-c}
    \end{subfigure}
  \caption{Diversity of retrieval datasets.}
  \label{fig:diversity}
\end{figure}

\subsection{Details of retrieved datasets}
Our approach is compatible with various retrieval datasets. Table 5 in the manuscript summarizes the size and performance of different retrieval sources. Notably, our model achieves the best results when using a professional photography retrieval set, such as the top 55,000 highest-rated images from AVA. The diversity of this AVA retrieval set is illustrated in \Cref{fig:diversity}, where we analyze the distribution of aesthetic scores, layout types, and semantic tags. As shown in \Cref{fig:diversity}, our retrieval dataset is diverse in semantic content, covers a wide range of layouts, and maintains high aesthetic quality.

\section{Text embedding}\label{sec:text}
\subsection{Feature extraction of text}
We use BLIP~\cite{li2022blip} to extract text embeddings for multi-modal fusion, as it is an open-source model freely available for use with new test sets. Although GPT-4 can produce more precise descriptions, its cost for processing test images limits its practicality for our approach. To maintain consistency between the training and testing phases, we rely on BLIP to generate text descriptions for multi-modal embedding fusion.

Additionally, we generate GPT-based text pairs specifically for outpainting purposes, used solely in creating images for our CAD dataset. These GPT-generated text pairs will also be released upon acceptance.

\subsection{Rationale of using text embedding}
Our work is motivated by the observation that language naturally highlights the most salient parts of an image, guiding the model in identifying key objects or regions. This approach mirrors human behavior, which focuses on the most important areas when viewing an image. While the aesthetic rules derived from retrieved images plays a crucial role in generating high-quality crops, incorporating text embeddings into the image embeddings provides marginal performance improvements, as summarized in Table 6 of manuscript.

\section{Layout features} 
\label{sec:layout}
Following previous work~\cite{ko2024semantic}, we assume that layout features can be characterized by patterns of layout combinations. We demonstrate that our extracted features effectively capture the line compositions of retrieved images, highlighting the advantages of our method over existing saliency-based approaches, particularly in complex scenarios.

\subsection{Visualization of SAM-extracted features}
\begin{figure}[!t]
  \centering
  \includegraphics[width=\linewidth]{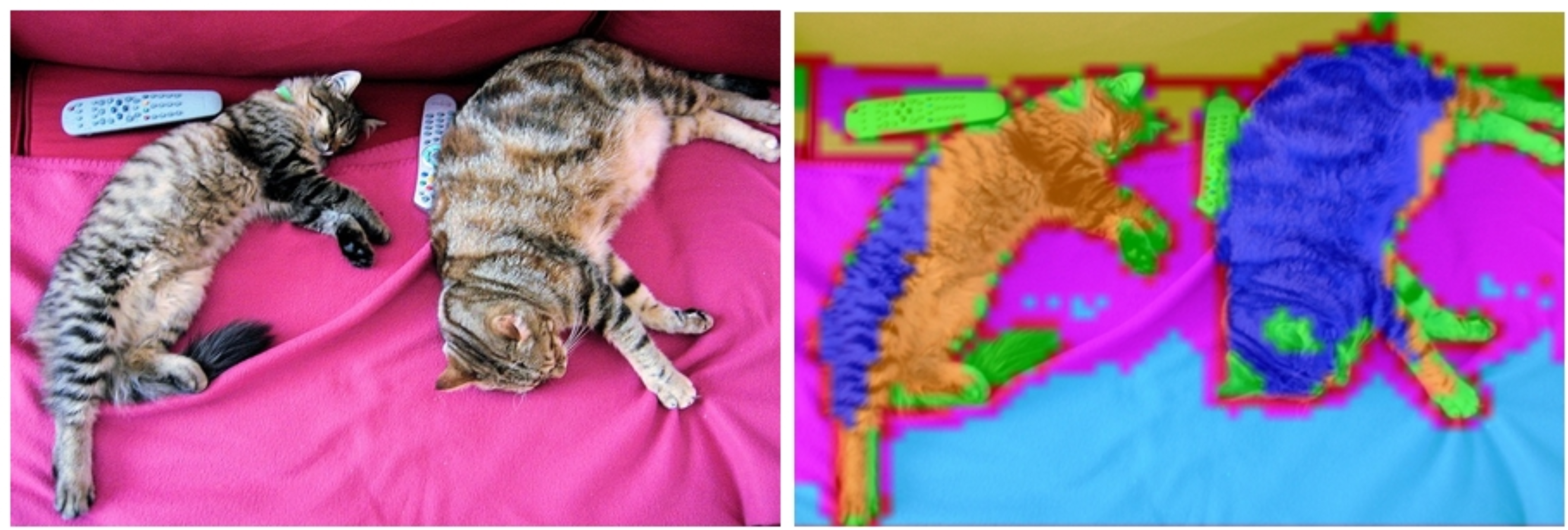}
  \caption{Visualization of SAM-extracted features (K-means clustering)~\cite{vs2024possam}.}
  \label{fig:sam}
\end{figure}

Instead of directly extracting geometric masks, we use the SAM encoder to obtain line and layout composition features from the query image, which are highly correlated with the geometric mask, as shown in \Cref{fig:sam}. We then treat the extracted layout (line combinations) as aesthetic guidelines and fuse them with the image embedding.

\begin{figure}[!t]
  \centering
  \includegraphics[width=\linewidth]{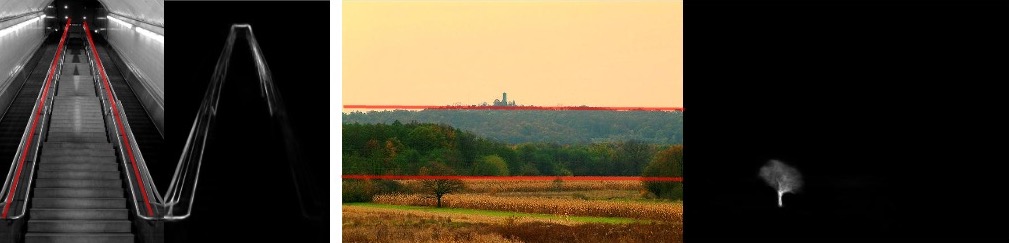}
  \caption{Saliency visualization in complex scenarios. We extract the saliency map following~\cite{horita2024retrieval}.}
  \label{fig:saliency}
\end{figure}

\subsection{Comparison to rule-based methods} 
Existing rule-based methods~\cite{hoh2023salient,zhang2018detecting,horita2024retrieval} focus on detecting salient objects, making them well-suited for images with simple, center compositions featuring a single prominent object. However, as shown in \Cref{fig:saliency}, these methods struggle with more complex scenes where no clear salient object exists. In contrast, our approach evaluates the overall layout composition by analyzing line structures. By retrieving professional images with similar line compositions as references, our method more effectively captures complex image layouts.

\section{Discussion}\label{sec:discuss}
\textbf{Limitations:} Our work has two primary limitations. First, the metrics used to evaluate aesthetic quality could be improved, as subjective annotations may not fully reflect the true aesthetic quality of the cropped areas. Second, we have not explored user control in the cropping process. In real-world applications, incorporating user-specific composition preferences could enable more personalized cropping styles.

\noindent\textbf{Future work:} We propose two promising directions for extending our work to more diverse scenarios. First, retrieved images could be used to guide the image generation process, allowing for finer control and enabling the creation of compositions with improved aesthetic quality. Second, the semantic similarity of the retrieved images makes them well-suited for segmentation tasks. Leveraging these images as references could improve fine-grained segmentation and help mitigate challenges associated with data scarcity.

\end{document}